\pdfoutput=1
\documentclass[11pt]{article}

\usepackage[final]{acl}

\usepackage{times}
\usepackage{latexsym}
\usepackage{booktabs}
\usepackage{multicol}
\usepackage{multirow}
\usepackage{graphicx}
\usepackage{tabularx}
\usepackage{hyperref}
\usepackage{listings}
\usepackage{array}
\usepackage{float}
\usepackage{xcolor}
\usepackage{enumitem}
\usepackage{longtable}
\usepackage[T1]{fontenc}
\usepackage[utf8]{inputenc}

\usepackage{microtype}

\usepackage{inconsolata}

\usepackage{graphicx}

\title{Tree of Problems: Improving structured problem solving\\ with compositionality}

\author{
  \textbf{Armel Zebaze},
  \textbf{Beno\^\i t Sagot},
  \textbf{Rachel Bawden}
\\
  Inria, Paris, France
\\
\texttt{\{firstname.lastname\}@inria.fr}
}

\begin{document}
\maketitle
\begin{abstract}
Large Language Models (LLMs) have demonstrated remarkable performance across multiple tasks through in-context learning. %Although consistent improvements are observed as models become larger and are trained on more data, they still face challenges with difficult tasks. 
For complex reasoning tasks that require step-by-step thinking, Chain-of-Thought (CoT) prompting has given impressive results, especially when combined with self-consistency. Nonetheless, some tasks remain particularly difficult for LLMs to solve. Tree of Thoughts (ToT) and Graph of Thoughts (GoT) emerged as alternatives, dividing the complex problem into paths of subproblems. In this paper, we propose Tree of Problems (ToP), a simpler version of ToT, which we hypothesise can work better for complex tasks that can be divided into identical subtasks. %how to use problem structure and compositionality to extend the range of task complexity LLMs can handle. 
Our empirical results show that our approach outperforms ToT and GoT, and in addition performs better than CoT on complex reasoning tasks. % breaking down a complex problem into a tree of related subproblems and integrating their solutions significantly enhances the accuracy of LLMs, outperforming strategies such as Tree of Thoughts and Graph of Thoughts.
All code for this paper is publicly available here: \url{https://github.com/ArmelRandy/tree-of-problems}.

\end{abstract}

\begin{figure*}[!ht]
  \center%\includegraphics[width=\columnwidth]{example-image-golden}
  \includegraphics[width=0.9\textwidth]{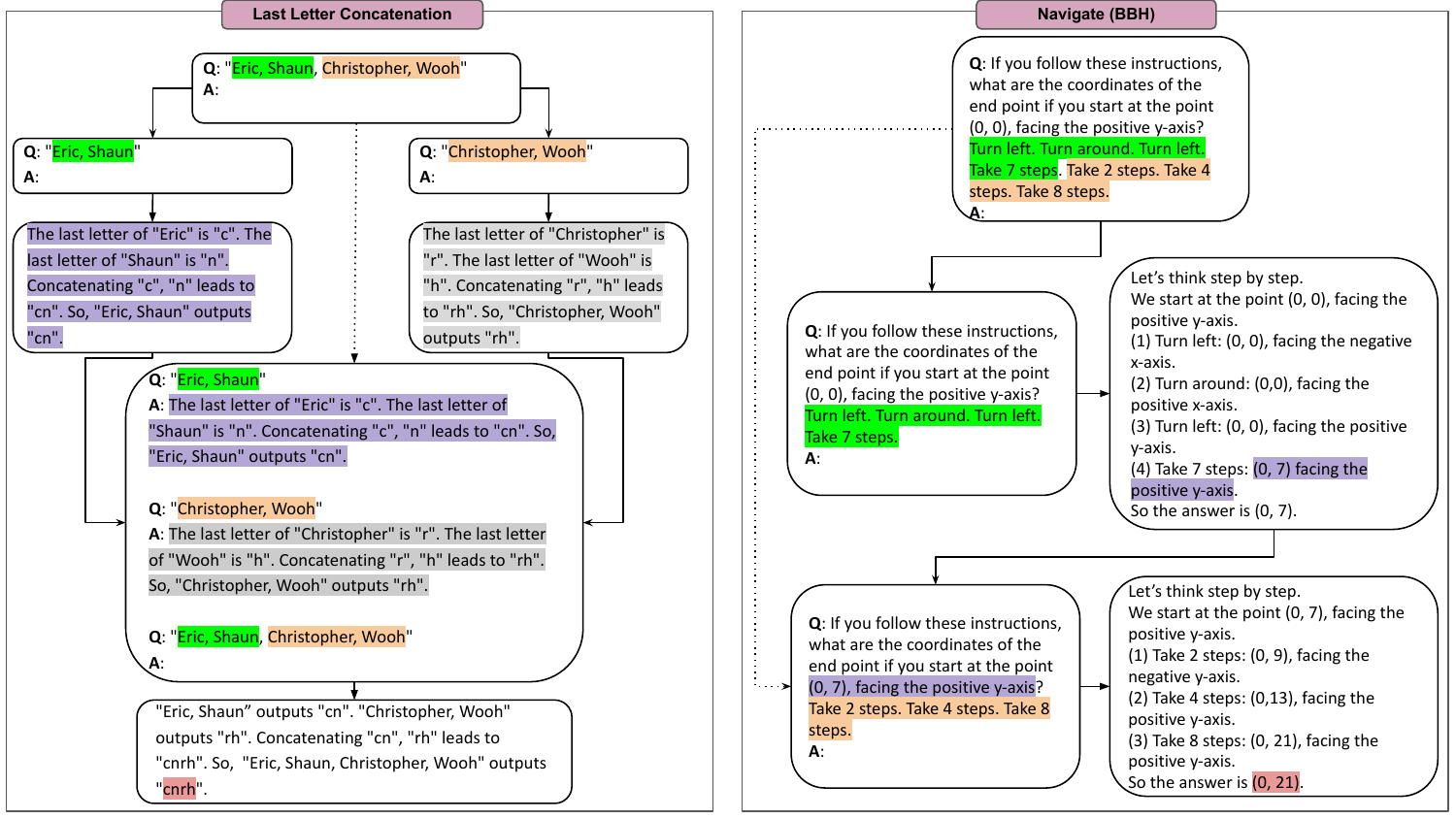}
  \caption{Overview of the Tree of Problems (ToP) framework for two tasks. On the left (a canonical task consisting of independent subproblems organised in a tree structure), the task is to concatenate the last letters of a list of names, accomplished by breaking the list in two, finding their solutions, and recombining them. On the right (an extension of the canonical structure to handle sequential tasks), the task is to determine the final position of an object after a series of steps. We first find its position after half of the steps, and then determine the final position by tracing the object through the remaining steps. See Section~\ref{sec:method} for a description of ToP.}
  \label{fig:fig1}
\end{figure*}

\section{Introduction}
In-Context Learning (ICL)~\citep{NEURIPS2020_1457c0d6} is the ability of Large Language Models (LLMs) to perform a task with the help of a few demonstrations within their context. It is widely used to evaluate LLMs on various tasks. These models, whose number of parameters and training corpus size has increased \iffalse often beyond power laws\fi{} massively over recent years, keep pushing the state of the art on a wide range of natural language tasks~\citep{anil2023palm, touvron2023llama, gemmateam2024gemma}. However, they still struggle to perform complex tasks, notably those requiring multiple reasoning steps~\citep{hendrycks2021measuring, hendrycksmath2021, suzgun-etal-2023-challenging}. Recently, Chain-of-Thought (CoT) prompting \citep{NEURIPS2022_9d560961, kojima2022large} has greatly helped to enhance reasoning abilities of LLMs by helping them to mimic step-by-step reasoning.
%Moreover, prompting a LLM to generate a long reasoning, even step-by-step, make it prone to errors that jeopardize its consistency which results into incorrect answers. 
However, CoT implicitly requires the model to generalize beyond the cases seen in its prompt, which often leads to poor out-of-domain performance~\citep{zhou2023leasttomost}. Applying CoT with self-consistency~\citep{wang2023selfconsistency} drives the model to explore multiple reasoning paths and to choose the most consistent answer, usually yielding better performance, but helping only marginally with out-of-distribution generalization. Moreover, solving complex problems involves understanding their underlying structure; this can help to avoid lengthy CoTs that are\iffalse considerably\fi{} prone to reasoning errors.

In this paper, we propose to tackle complex problem-solving and out-of-distribution generalization by dividing complex tasks into a series of simpler sub-tasks. We draw inspiration from %algorithm design, with 
techniques such as dynamic programming and divide and conquer in order to efficiently guide LLMs through complex problem solving. 
Such problems have previously been tackled using approaches adding structure to CoT, such as Tree of Thoughts (ToT) \citep{yao2023tree} and Graph of Thoughts (GoT)
\citep{Besta_Blach_Kubicek_Gerstenberger_Podstawski_Gianinazzi_Gajda_Lehmann_Niewiadomski_Nyczyk_Hoefler_2024}, which consist in sampling diverse reasoning paths (where path states represent subproblems) and finding the optimal path.
We argue that for a subset of complex reasoning problems, where an instance can be decomposed into multiple analogous subinstances, ToT and GoT are overly complex, and the tasks can be better solved by a simpler approach.
%We argue that for a subset of complex reasoning problems, where subproblems can be extracted and solved independently of each other, ToT and GoT are overly complex, and the tasks can be better solved by a simpler approach.
%
This simpler approach, which we name Tree of Problems (ToP) consists in building a tree structure, where each node represents a problem instance similar to the main instance. The deepest instances, which correspond to atomic problems, are solved first with CoT prompting and the internal nodes are recursively solved by merging their children's solutions. Figure~\ref{fig:fig1} illustrates our method on the tasks of Last Letter Concatenation and Navigate from the BIG-Bench Hard benchmark~\citep{suzgun-etal-2023-challenging}. %We start by  We name this process of building a tree of problem instances and recursively solving them in a bottom-up approach the ``Tree of Problems'' framework (ToP).

We conduct a comprehensive evaluation on several LLMs, including GPT-3.5, on multiple hard tasks. We find that ToP\iffalse greatly\fi{} improves LLMs' problem solving abilities on structured tasks outperforming CoT, ToT and GoT by a large margin.

%\begin{figure}[t]
  %\includegraphics[width=\columnwidth]{example-image-golden}
%  \includegraphics[width=\columnwidth]{figures/fig2.pdf}
%  \caption{A figure with a caption that runs for more than one line.
%    Example image is usually available through the \texttt{mwe} package
%    without even mentioning it in the preamble.}
%  \label{fig:fig1}
%\end{figure}

\section{Related Work}
%\paragraph{Reasoning in Language Models.}
CoT prompting was proposed to enhance reasoning by incorporating step-by-step logic into few-shot prompt demonstrations \citep{NEURIPS2022_9d560961}. It showed significant improvement over standard input-output (IO) prompting across various mathematical and symbolic reasoning benchmarks. Building on this, \citet{kojima2022large} and \citet{wang-etal-2023-plan} \textit{inter alia} demonstrated that zero-shot CoT could be achieved %in a zero-shot setting 
by using reasoning-inducing words at the end of the zero-shot prompt. Other works showed that wisely designing the CoT demonstrations could yield further improvements \citep{zhang2023automatic, fu2022complexity}. CoT Self-Consistency (CoT-SC; \citealt{wang2023selfconsistency}) improved on CoT by sampling diverse reasoning steps and selecting the most consistent answer after marginalizing over the reasoning paths. Our research also builds on the %extensive 
body of work addressing problem-solving through compositionality, which involves teaching LLMs to tackle complex problems by breaking them down into a series of subproblems and recursively solving them to derive the final answer, e.g.~Least-to-Most \citep{zhou2023leasttomost}, decomposed \citep{khot2023decomposed} and successive \citep{dua-etal-2022-successive} prompting. While these works align with our approach through their use of problem decomposition, we focus on breaking a main task into multiple similar subtasks, solvable using the same prompt. Moreover, our approach uses a tree structure that allows for greater flexibility and coverage in problem solving. The most closely related approaches are Tree of Thoughts (ToT) \citep{yao2023tree} and Graph of Thoughts (GoT) \citep{Besta_Blach_Kubicek_Gerstenberger_Podstawski_Gianinazzi_Gajda_Lehmann_Niewiadomski_Nyczyk_Hoefler_2024}. ToT builds on the idea of sampling diverse reasoning paths but redefines problem solving as a search over a thought space, where states represent partial solutions. GoT extends ToT by including thought aggregation, which is analogous to our merge operation and by allowing refining~\citep{madaan2023selfrefine}. While a ``thought'' represents a general reasoning step in their approach, we focus on reasoning through subproblems. We do not perform a search over a tree of thoughts, nor do we score or refine (improve) our tree nodes. Instead, each node in the tree of problems is directly relevant to solving the problem, and their bottom-up recombination produces the final solution. ToP is therefore a simpler and more cost-effective alternative to ToT and GoT.

\section{Our method}\label{sec:method}
Solving a complex problem often requires reasoning, partly explaining the success of CoT prompting for such problems. Reasoning involves understanding a problem's structure and design. This aspect is frequently overlooked in CoT because incorporating it can be challenging. \iffalse When a task instance is simple, understanding its structure is less crucial. However, as task complexity increases, a deeper and broader understanding becomes necessary.\fi{} Our method addresses this by constructing a tree of simpler, closely related subproblems to solve a more complex problem. We hypothesize that the capability of an LLM to solve simple instances can be extended to more complex ones. The ability of an LLM to solve a complex instance therefore lies in how accurately it can solve simpler ones and then combine their answers. The main class of problems we aim to tackle are complex problems that are divisible into independent subproblems resembling the initial one (we refer to these as canonical tasks). However, we also experiment with relaxing the independency constraint in order to tackle sequential tasks, which require finding the final state of a system after a series of independent processing steps (See the right of Figure~\ref{fig:fig1}). Our method relies on the following components:

\begin{itemize}[noitemsep, topsep=0pt, leftmargin=*]
    \item \textbf{A decomposer} divides a problem instance into a series of smaller related instances, algorithmically or via few-shot prompting with a \verb|divide_prompt|. We recursively build a tree of problems (nodes) considering 2 parameters: the \textit{breadth} (the number of children of each internal node) and the \textit{depth} of the tree, directly related to the granularity of the atomic subproblems. The root of the tree is the main problem. In this paper, ToP~($b$, $d$) refers to using breadth $b$ and depth $d$.
    \item \textbf{A solver} is used to do the task of interest, namely the simplest instances obtained after decomposition (in our case an LLM with a task-specific \verb|solve_prompt|).
    \item \textbf{A merger} receives the solved subproblems (problem statement and solution) at level \textit{k} to build and solve the problem at level $k-1$. It uses a specific \verb|merge_prompt| to get the LLM to learn to combine the subproblems' solutions into the parent solution. As opposed to L2M, the prompt to get the solution of a problem at a level $\textit{k}$ only depends on the directly connected problems (at the level $k+1$). 
\end{itemize}
The workflow can be described as follows: The decomposer builds the tree of problems, the solver addresses the subproblems at the tree's leaves, and the merger recursively derives each node's solution by combining its children's solutions in a bottom-up approach. The total number of inference calls %required by ToP 
(omitting the cost of problem decomposition) is equal to the number of nodes in the tree structure. 

In addition to canonical tasks with a classic tree structure (see the left of Figure~\ref{fig:fig1}), ToP can also be used for sequential tasks, where a given subproblem needs the result of a previous subproblem as an input (see the right of Figure~\ref{fig:fig1}). 
Our standard ToP paradigm described above can be used to solve such problems by setting the breadth to 1.
This has the effect that the problem is decomposed into a sequence of $n$ subproblems organised as hierarchy of depth $n$. When solving the $(k+1)$-th subproblem, the solver will have access to its child subproblem's result, i.e.~the result of subproblem $k$, thereby accounting for the sequentiality of the decomposition. 
%A subproblem involves identifying the system's state after a subset of the steps (e.g.~half of the total steps), with the resulting solution defining the subsequent problem (state) to which to apply the remaining steps. In this setup, the breadth should be set to 1 and the tree becomes a path graph because the formulation of the problem to and the level $k$ requires the answer at the level $k + 1$ (See Figure~\ref{fig:fig1}).
%
The LLM is no longer required to merge subproblems' solutions; it is directly fed with a new problem formulation automatically computed using the corresponding child's solution. The final solution is obtained by solving the last subproblem, and so the main problem instance (root node) does not influence the inference cost.

For both tasks, all problems at the same level of the tree are solved in parallel to promote efficiency. We further detail the method with more examples in Appendix~\ref{app:clarifications}.

\section{Experiments}

%\subsection{Experimental Setup}

We first compare ToP to ToT and GoT to test our hypothesis that our simpler approach is more adapted to canonical tasks. %(complex problems that are divisible into independent subproblems resembling the initial one). 
We do this using the GoT tasks proposed by \citet{Besta_Blach_Kubicek_Gerstenberger_Podstawski_Gianinazzi_Gajda_Lehmann_Niewiadomski_Nyczyk_Hoefler_2024}. We then show that ToP is more effective in comparison to IO (direct input-output) and CoT prompting across a wider ranger of canonical tasks, namely Last Letter Concatenation~\citep{NEURIPS2022_9d560961} and 5 BIG-Bench Hard~\citep{srivastava2023beyond} tasks fitting the description. Finally, we test ToP on sequential tasks.

\subsection{Datasets}\label{datasets}
\paragraph{GoT tasks.} \citet{Besta_Blach_Kubicek_Gerstenberger_Podstawski_Gianinazzi_Gajda_Lehmann_Niewiadomski_Nyczyk_Hoefler_2024} compared GoT to ToT, IO, and CoT prompting on three tasks (each with 100 examples): (i)~\textit{Sorting}, which involves arranging a list of 32 numbers ranging from 0 to 9 (both inclusive) in order, (ii)~\textit{Set Intersection}, which involves finding the common elements between two sets, each containing 32 elements and (iii)~\textit{Keyword Counting}, which involves identifying countries mentioned in a text and counting how many times each country appears. %Each of these tasks has 100 examples.

\paragraph{Symbolic Reasoning.} We use two toy tasks introduced by \citet{NEURIPS2022_9d560961} (each with 500 examples): (i)~\textit{Last Letter Concatenation}, where the LLM is tasked with recovering the concatenation of the last letters from a list of names and (ii)~\textit{Coin Flip}, which evaluates if the LLM can deduce the final state of a coin (heads or tails) after people either flip it or not. During evaluation, we consider various list lengths (4, 8 and 16) for the first task, and different numbers of people involved (4, 8 and 16) for the second. %Each of these tasks has 500 examples.

\paragraph{BIG-Bench Hard (BBH).} BBH consists of 23\iffalse challenging\fi{} BIG-Bench~\citep{srivastava2023beyond} tasks that have been shown to benefit from CoT~\citep{suzgun-etal-2023-challenging}. We use 8 tasks:\footnote{\iffalse Initially, \textit{Navigate} and \textit{Tracking Shuffled Objects} are multiple choice questions. We modify them to require the LLM to output respectively the coordinate of the final position and the final matching of objects. Additionally, we turn \textit{Hyperbaton} into a Yes/No task.\fi See Appendix~\ref{section:modification} for more details.} \textit{Boolean Expressions}, \textit{Hyperbaton}, \textit{Multi-Step Arithmetic Two}, \textit{Navigate}, \textit{Object Counting}, \textit{Tracking Shuffled Objects (3, 5, 7)}, \textit{Web of Lies} and \textit{Word Sorting}.

\subsection{Language models and prompts}
We experiment with \texttt{gpt-3.5-turbo} and \texttt{gpt-3.5-turbo-instruct}.\footnote{More results and analysis for LLaMA (different model versions and sizes) are provided in Appendices~\ref{appendix:scaling} and~\ref{appendix:analysis}.
} For the \verb|solve_prompts|, we use the CoT prompts\footnote{We report some results with IO in Appendix~\ref{appendix:robustness}.} of~\citet{suzgun-etal-2023-challenging} on BBH tasks, with minor changes.
\iffalse \footnote{For \textit{Hyperbaton}, \textit{Navigate} and \textit{Tracking Shuffled Objects.}}\fi The CoT prompts for \textit{Symbolic Reasoning} are inspired by those in \citep{NEURIPS2022_9d560961}, which contain 8 examples of 2-letters or 2-flips and those for \textit{GoT tasks} are the same as in \citet{Besta_Blach_Kubicek_Gerstenberger_Podstawski_Gianinazzi_Gajda_Lehmann_Niewiadomski_Nyczyk_Hoefler_2024}. We report some implementation details in Appendix~\ref{appendix:details} and Appendix~\ref{app:prompts}.

%and \href{https://huggingface.co/casperhansen/llama-3-70b-instruct-awq}{LLaMA 3 70B Instruct} (instead of LLaMA 3 70B
%\begin{figure*}[t]
  %\includegraphics[width=0.48\linewidth]{example-image-a} \hfill
  %\includegraphics[width=0.48\linewidth]{example-image-b}
%  \includegraphics[width=0.48\linewidth]{figures/l2m.pdf} \hfill
%  \includegraphics[width=0.48\linewidth]{figures/top.pdf}
%  \caption {A minimal working example to demonstrate how to place
%    two images side-by-side.}
%\end{figure*}

\subsection{Main results}

%\paragraph{GoT tasks.} These tasks fit perfectly within the ToP framework. A list can be recursively sorted similarly to merge sort. Finding the intersection of two sets can be done by splitting each set into two subsets, and counting occurrences in a text can be handled by processing it block by block.
\paragraph{GoT tasks.}
Table~\ref{tab:topgottot} compares our results on the GoT tasks with those obtained by rerunning the CoT, ToT and GoT approaches from \citep{Besta_Blach_Kubicek_Gerstenberger_Podstawski_Gianinazzi_Gajda_Lehmann_Niewiadomski_Nyczyk_Hoefler_2024}. More precisely, we use the highest accuracy achieved with ToT and GoT on each task with \texttt{gpt-3.5-turbo-0125}\iffalse as of June 2024\fi{}. For \textit{Sorting}, we intuitively choose $b = 2$ as in merge sort and $d = 2$ for performance. We use the same $b$ for \textit{Keyword Counting}, with $d = 4$ to get simple atomic instances. In \textit{Set Intersection}, we use $b = 4$ because each set is divided into two disjoint subsets, resulting in four pairs of subsets (one pair per subproblem). Such a large breadth was sufficient to produce simple atomic problems, so we used $d = 1$. ToP outperforms ToT and GoT by a large margin on \textit{sorting} with an absolute improvement of 40\% over GoT. Similarly, ToP outperforms GoT by 19\% and 5\% respectively on \textit{Set Intersection} and \textit{Keyword Counting}.

\begin{table}[ht]
\small
\begin{center}
\resizebox{0.5\textwidth}{!}{
\begin{tabular}{lrrrr}
\toprule
\multirow{2}{*}{\bf GoT Tasks}  & \multicolumn{4}{c}{\texttt{gpt-3.5-turbo}} \\
\cmidrule{2-5}
{} & CoT & ToT (best) & GoT (best) & ToP (ours) \\
\midrule
\textit{Sorting} & 0.02 & 0.17 & 0.28 & \bf 0.68 \\
\textit{Set Intersection} & 0.07 & 0.25 & 0.46 & \bf 0.65 \\
\textit{Keyword Counting} & 0.00 & 0.00 & 0.26 & \bf 0.31 \\
\bottomrule
\end{tabular}
}
\end{center}
\caption{Results on 3 tasks from \citep{Besta_Blach_Kubicek_Gerstenberger_Podstawski_Gianinazzi_Gajda_Lehmann_Niewiadomski_Nyczyk_Hoefler_2024}\iffalse with \texttt{gpt-3.5-turbo}\fi. In all results tables, best results are highlighted in bold.}
\label{tab:topgottot}
\end{table}

\paragraph{Last Letter Concatenation.}
We consider ToP~(2,\,1). Subproblems are obtained by dividing the main list into \textit{b}~$=2$ lists of equal length.
\begin{table}[ht]
\centering\tiny
\begin{center}
\resizebox{0.5\textwidth}{!}{
%\resizebox{\columnwidth}{!}{
%\resizebox{!}{2cm}{
\begin{tabular}{lrrrr}
\toprule
\multirow{2}{*}{\bf Last Letter Concatenation}  & \multicolumn{3}{c}{\texttt{gpt-3.5-turbo-instruct}} \\
\cmidrule{2-4}
{} & IO & CoT & ToP (ours) \\
\midrule
%\multicolumn{3}{l}{\textit{Last Letter Concatenation}} \\
%\textit{Last Letter Concatenation}\\
%\midrule
\textit{Four} & 0.032 & 0.900 & \bf 0.990 \\
\textit{Eight} & 0.000 & 0.662 & \bf 0.854 \\
\textit{Sixteen} & 0.000 & 0.252 & \bf 0.444\\
\bottomrule
\end{tabular}
}
\end{center}
\caption{Results on Symbolic Reasoning tasks.}
\label{tab:symbolic}
\end{table}

\paragraph{Comparison to Least-to-Most Prompting and CoT with Self-consistency.} Least-to-most (L2M) prompting has also been successfully applied to Last Letter Concatenation~\citep{zhou2023leasttomost}. Given a list of $L$ names, L2M requires $L - 1$ inference calls, the first to concatenate the first 2 last letters and the $L - 2$ other to add the remaining last letters one after the other. Following~\citet{huang2024large}, we provide a fair comparison of L2M to ToP by adapting ToP's tree structure to require the same number of inference calls as L2M. This is done by using trees of breadth $2$ and depth $log_2(L) - 1$ for lists of length $L$. We compare ToP to L2M as well as CoT self-consistency with $L$ reasoning paths. The results (Table~\ref{tab:symbolic_gpt}) show that for L = 4 or L = 8, ToP (breadth = 2, depth = 1) achieves comparable performance to L2M while requiring half as many inference calls. When the number of inference calls is matched between the two methods, ToP demonstrates superior performance in all scenarios. CoT-SC lags behind both L2M and ToP.

\begin{table}[ht]
\small
\begin{center}
\resizebox{0.5\textwidth}{!}{
\begin{tabular}{lrrrrr}
\toprule
%\multirow{2}{*}{\bf Symbolic Reasoning}  & \multicolumn{4}{c}{\texttt{gpt-3.5-turbo-instruct}} \\
\multirow{2}{*}{\bf Last Letter Concatenation}  & \multicolumn{4}{c}{\texttt{gpt-3.5-turbo-instruct}} \\
\cmidrule{2-5}
{} & CoT-SC & L2M & ToP & ToP (match) \\
\midrule
%\textit{Last Letter Concatenation} \\
%\midrule
\textit{Four} & 0.908 & 0.988 & 0.990 & \bf 0.990 & \\
\textit{Eight} & 0.574 & 0.870 & 0.854 & \bf 0.932 & \\
\textit{Sixteen} & 0.116 & 0.742 & 0.444 & \bf 0.858 & \\
\bottomrule
\end{tabular}
}
\end{center}
\caption{Comparison of ToP to L2M and CoT-SC for Last Letter Concatenation. ToP (match) refers to ToP with the same number of inference calls as L2M.}
\label{tab:symbolic_gpt}
\end{table}

\begin{figure}[t]
  \includegraphics[width=0.95\columnwidth]{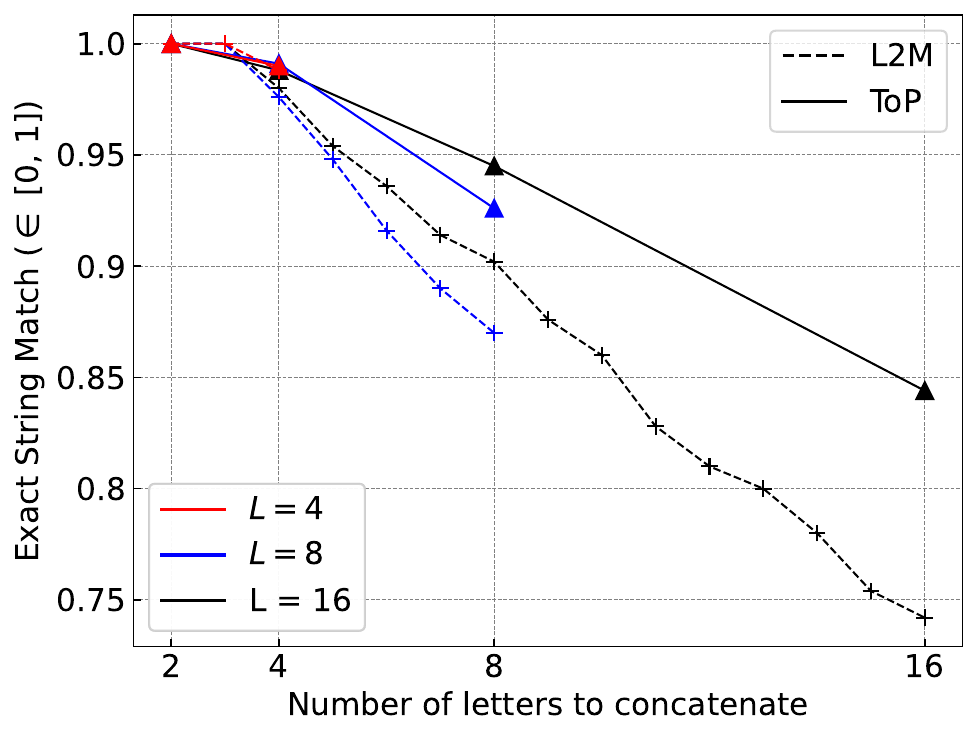}
  \caption {Per-level accuracy of Least to Most prompting and ToP (match) for \textit{Last Letter Concatenation}.}
  \label{fig:l2mvstop}
\end{figure}

Moreover, since L2M is similar to ToP~(1, $L-1$), we compare its accuracy to ToP (match) at each level of the tree. As illustrated in Figure~\ref{fig:l2mvstop}, both methods start with a perfect score that gradually decreases as they approach the task's resolution. ToP (match) consistently outperforms L2M at each step across all three settings.

\subsection{Complementary results}
We have successfully applied ToP to problems that can be divided into multiple independent instances. In this section, we report additional results for more such tasks and sequential tasks. 

\subsubsection{Canonical BBH tasks}
BBH tasks such as \textit{Boolean Expressions}, \textit{Hyperbaton}, \textit{Multistep Arithmetic Two}, \textit{Object Counting}, and \textit{Word Sorting} can be decomposed into multiple independent instances, whose solutions are later combined. They therefore correspond to canonical ToP tasks. We apply ToP~(2, 1) to them and report results in Table~\ref{tab:bbh_canonical}. 
ToP yields an absolute improvement over CoT of 21.2\% on \textit{Word Sorting} and 9.8\% on \textit{Hyperbaton}. However, it is slightly worse than CoT on \textit{Boolean Expressions}, \textit{Multistep Arithmetic Two} and \textit{Object Counting} with an average deterioration of 3.6\% on the 3 tasks. We attribute this loss of accuracy to reasoning inconsistencies and we explore this in more detail in Appendix~\ref{appendix:failure_cases}.

\begin{table}[ht]
\centering\small
%\resizebox{0.5\textwidth}{!}{
\resizebox{\columnwidth}{!}{
%\resizebox{!}{2.5cm}{
\begin{tabular}{lrrrr}
\toprule
\multirow{2}{*}{}  & \multicolumn{3}{c}{\texttt{gpt-3.5-turbo-instruct}} \\
\cmidrule{2-4}
{} & IO & CoT & ToP \\
\midrule
\textit{Boolean Expressions} & 0.908 & \bf 0.924 &  0.896 \\
\textit{Hyperbaton} & 0.528 & 0.804 & \bf 0.902 \\
\textit{Multistep Arithmetic Two} & 0.032 & \bf 0.780 & 0.736 \\
\textit{Object Counting} & 0.412 & \bf 0.928 & 0.892 \\
\textit{Word Sorting} & \bf 0.837 & 0.619 & 0.831 \\
\bottomrule
\end{tabular}
}
\caption{Results on the canonical BBH tasks.}
\label{tab:bbh_canonical}
\end{table}

\subsubsection{Sequential tasks}
\textit{Coin Flip} is an example of a sequential task. Using ToP~(1, 2), the problem at the leaves is to find the state of the coin after going through the first half of the people. The final solution is obtained by determining how this state changes as the coin goes through the remaining half of the people. \textit{Navigate}, \textit{Tracking Shuffled Objects}, and \textit{Web of Lies} can be modeled in a similar way. ToP outperforms CoT on all tasks, as shown in Table~\ref{tab:sequential}. ToP reaches near perfect accuracy on \textit{Coin Flip} with 4 and 8 people. Moreover, it is more robust to the out-of-distribution setting than CoT, showing a lesser performance drop as the number of people increases. Compared to CoT, it yields an absolute improvement of 5.2\% on \textit{Web of Lies}, 5.9\% on average on \textit{Tracking Shuffled Objects} and 2\% on \textit{Navigate}.

\begin{table}[ht]
\centering\small
%\resizebox{0.5\textwidth}{!}{
\resizebox{\columnwidth}{!}{
%\resizebox{!}{2.5cm}{
\begin{tabular}{lrrrr}
\toprule
\multirow{2}{*}{}  & \multicolumn{3}{c}{\texttt{gpt-3.5-turbo-instruct}} \\
\cmidrule{2-4}
{} & IO & CoT & ToP \\
\midrule
\textit{Coin Flip} \\
\midrule
\textit{Four} & 0.512 & \bf 0.998 &  \bf 0.998\\
\textit{Eight} & 0.502 & 0.840 & \bf 0.998 \\
\textit{Sixteen} & 0.476 & 0.718 & \bf 0.756 \\
\midrule
\textit{BIG-Bench Hard} \\
\midrule
\textit{Navigate} & 0.204 & 0.864 & \bf 0.884\\
\textit{Tracking Shuffled Objects (3)} & 0.004 & \bf 0.536 &  0.524 \\
\textit{Tracking Shuffled Objects (5)} & 0.004 & 0.324 & \bf 0.440 \\
\textit{Tracking Shuffled Objects (7)} & 0.000 & 0.044 & \bf 0.118 \\
\textit{Web of Lies} & 0.528 & 0.920 & \bf 0.972\\
\bottomrule
\end{tabular}
}
\caption{Results on \textit{Coin Flip} and sequential BBH tasks.}
\label{tab:sequential}
\end{table}

\section*{Conclusion}
We use compositionality to grant the ability to solve complex and structured problems to LLMs via the Tree of Problems (ToP) framework. ToP is a simplification of the ToT framework, and involves decomposing complex tasks into identical subtasks. %stems from the necessity to understand the structure of problems to solve them as they grow in complexity.
%from various scales\footnote{See Appendix~\ref{appendix:scaling}}
Our experiments demonstrate that LLMs can benefit from ToP and solve certain complex problems better than ToT, GoT and L2M approaches and generalize better than with CoT.

\section*{Acknowledgements}
This work was partly funded  by the last two authors' chairs in the PRAIRIE institute funded by the French national agency ANR as part of the ``Investissements d'avenir'' programme under the reference ANR-19-P3IA-0001. The authors are grateful to the OPAL infrastructure from Université Côte d'Azur for providing resources and support. We thank Maciej Besta for answering our questions regarding Graph of Thoughts and Lydia Nishimwe for her helpful feedback.

%\newpage

\section*{Limitations}
\paragraph{Applicability of the Tree of Problems framework.} Although ToP is a powerful prompting strategy that significantly widens the range of tasks that LLMs can handle accurately; it is limited to problems which have a structure (mathematical tasks, algorithmic tasks etc.) that can be decomposed into analogous subproblems. The founding hypothesis of ToP is the fact that LLMs can solve simple instances of a task and this ability can be efficiently translated to more complex instances. %Problems such as open-ended question answering would require to make some changes to the framework as it currently stands in order to broaden its applicability. 
%However, doing so might break the founding assumption of ToP and thus make its performance unpredictable. Allowing more flexibility typically results in more sophisticated techniques such as tree-of-thoughts and graph-of-thoughts.

\paragraph{Reasoning consistency of LLMs.} %In addition to the modest size of the problems on which ToP is applicable, 
LLMs can surprisingly fail to be robust to minor changes in a problem formulation. They can fail to solve a problem closely related to another one that they are capable to solve. We note this as a typical failure case of ToP in Appendix~\ref{appendix:failure_cases} on \textit{Object Counting} and \textit{Multistep Arithmetic Two}.

%\section*{Acknowledgments}

%This document has been adapted
%by Steven Bethard, Ryan Cotterell and Rui Yan
%from the instructions for earlier ACL and NAACL proceedings, including those for
%ACL 2019 by Douwe Kiela and Ivan Vuli\'{c},
%NAACL 2019 by Stephanie Lukin and Alla Roskovskaya,
%ACL 2018 by Shay Cohen, Kevin Gimpel, and Wei Lu,
%NAACL 2018 by Margaret Mitchell and Stephanie Lukin,
%Bib\TeX{} suggestions for (NA)ACL 2017/2018 from Jason Eisner,
%ACL 2017 by Dan Gildea and Min-Yen Kan,
%NAACL 2017 by Margaret Mitchell,
%ACL 2012 by Maggie Li and Michael White,
%ACL 2010 by Jing-Shin Chang and Philipp Koehn,
%ACL 2008 by Johanna D. Moore, Simone Teufel, James Allan, and Sadaoki Furui,
%ACL 2005 by Hwee Tou Ng and Kemal Oflazer,
%ACL 2002 by Eugene Charniak and Dekang Lin,
%and earlier ACL and EACL formats written by several people, including
%John Chen, Henry S. Thompson and Donald Walker.
%Additional elements were taken from the formatting instructions of the \emph{International Joint Conference on Artificial Intelligence} and the \emph{Conference on Computer Vision and Pattern Recognition}.

% Bibliography entries for the entire Anthology, followed by custom entries
\bibliography{anthology, bib}

\begin{thebibliography}{22}
\providecommand{\natexlab}[1]{#1}

\bibitem[{Anil et~al.(2023)Anil, Dai, Firat, Johnson, Lepikhin, Passos, Shakeri, Taropa, Bailey, Chen, Chu, Clark, Shafey, Huang, Meier-Hellstern, Mishra, Moreira, Omernick, Robinson, Ruder, Tay, Xiao, Xu, Zhang, Abrego, Ahn, Austin, Barham, Botha, Bradbury, Brahma, Brooks, Catasta, Cheng, Cherry, Choquette-Choo, Chowdhery, Crepy, Dave, Dehghani, Dev, Devlin, Díaz, Du, Dyer, Feinberg, Feng, Fienber, Freitag, Garcia, Gehrmann, Gonzalez, Gur-Ari, Hand, Hashemi, Hou, Howland, Hu, Hui, Hurwitz, Isard, Ittycheriah, Jagielski, Jia, Kenealy, Krikun, Kudugunta, Lan, Lee, Lee, Li, Li, Li, Li, Li, Lim, Lin, Liu, Liu, Maggioni, Mahendru, Maynez, Misra, Moussalem, Nado, Nham, Ni, Nystrom, Parrish, Pellat, Polacek, Polozov, Pope, Qiao, Reif, Richter, Riley, Ros, Roy, Saeta, Samuel, Shelby, Slone, Smilkov, So, Sohn, Tokumine, Valter, Vasudevan, Vodrahalli, Wang, Wang, Wang, Wang, Wieting, Wu, Xu, Xu, Xue, Yin, Yu, Zhang, Zheng, Zheng, Zhou, Zhou, Petrov, and Wu}]{anil2023palm}
Rohan Anil, Andrew~M. Dai, Orhan Firat, Melvin Johnson, Dmitry Lepikhin, Alexandre Passos, Siamak Shakeri, Emanuel Taropa, Paige Bailey, Zhifeng Chen, Eric Chu, Jonathan~H. Clark, Laurent~El Shafey, Yanping Huang, Kathy Meier-Hellstern, Gaurav Mishra, Erica Moreira, Mark Omernick, Kevin Robinson, Sebastian Ruder, Yi~Tay, Kefan Xiao, Yuanzhong Xu, Yujing Zhang, Gustavo~Hernandez Abrego, Junwhan Ahn, Jacob Austin, Paul Barham, Jan Botha, James Bradbury, Siddhartha Brahma, Kevin Brooks, Michele Catasta, Yong Cheng, Colin Cherry, Christopher~A. Choquette-Choo, Aakanksha Chowdhery, Clément Crepy, Shachi Dave, Mostafa Dehghani, Sunipa Dev, Jacob Devlin, Mark Díaz, Nan Du, Ethan Dyer, Vlad Feinberg, Fangxiaoyu Feng, Vlad Fienber, Markus Freitag, Xavier Garcia, Sebastian Gehrmann, Lucas Gonzalez, Guy Gur-Ari, Steven Hand, Hadi Hashemi, Le~Hou, Joshua Howland, Andrea Hu, Jeffrey Hui, Jeremy Hurwitz, Michael Isard, Abe Ittycheriah, Matthew Jagielski, Wenhao Jia, Kathleen Kenealy, Maxim Krikun, Sneha Kudugunta, Chang
  Lan, Katherine Lee, Benjamin Lee, Eric Li, Music Li, Wei Li, YaGuang Li, Jian Li, Hyeontaek Lim, Hanzhao Lin, Zhongtao Liu, Frederick Liu, Marcello Maggioni, Aroma Mahendru, Joshua Maynez, Vedant Misra, Maysam Moussalem, Zachary Nado, John Nham, Eric Ni, Andrew Nystrom, Alicia Parrish, Marie Pellat, Martin Polacek, Alex Polozov, Reiner Pope, Siyuan Qiao, Emily Reif, Bryan Richter, Parker Riley, Alex~Castro Ros, Aurko Roy, Brennan Saeta, Rajkumar Samuel, Renee Shelby, Ambrose Slone, Daniel Smilkov, David~R. So, Daniel Sohn, Simon Tokumine, Dasha Valter, Vijay Vasudevan, Kiran Vodrahalli, Xuezhi Wang, Pidong Wang, Zirui Wang, Tao Wang, John Wieting, Yuhuai Wu, Kelvin Xu, Yunhan Xu, Linting Xue, Pengcheng Yin, Jiahui Yu, Qiao Zhang, Steven Zheng, Ce~Zheng, Weikang Zhou, Denny Zhou, Slav Petrov, and Yonghui Wu. 2023.
\newblock \href {https://arxiv.org/abs/2305.10403} {Palm 2 technical report}.
\newblock \emph{Preprint}, arXiv:2305.10403.

\bibitem[{Besta et~al.(2024)Besta, Blach, Kubicek, Gerstenberger, Podstawski, Gianinazzi, Gajda, Lehmann, Niewiadomski, Nyczyk, and Hoefler}]{Besta_Blach_Kubicek_Gerstenberger_Podstawski_Gianinazzi_Gajda_Lehmann_Niewiadomski_Nyczyk_Hoefler_2024}
Maciej Besta, Nils Blach, Ales Kubicek, Robert Gerstenberger, Michal Podstawski, Lukas Gianinazzi, Joanna Gajda, Tomasz Lehmann, Hubert Niewiadomski, Piotr Nyczyk, and Torsten Hoefler. 2024.
\newblock \href {https://doi.org/10.1609/aaai.v38i16.29720} {Graph of thoughts: Solving elaborate problems with large language models}.
\newblock \emph{Proceedings of the AAAI Conference on Artificial Intelligence}, 38(16):17682--17690.

\bibitem[{Brown et~al.(2020)Brown, Mann, Ryder, Subbiah, Kaplan, Dhariwal, Neelakantan, Shyam, Sastry, Askell, Agarwal, Herbert-Voss, Krueger, Henighan, Child, Ramesh, Ziegler, Wu, Winter, Hesse, Chen, Sigler, Litwin, Gray, Chess, Clark, Berner, McCandlish, Radford, Sutskever, and Amodei}]{NEURIPS2020_1457c0d6}
Tom Brown, Benjamin Mann, Nick Ryder, Melanie Subbiah, Jared~D Kaplan, Prafulla Dhariwal, Arvind Neelakantan, Pranav Shyam, Girish Sastry, Amanda Askell, Sandhini Agarwal, Ariel Herbert-Voss, Gretchen Krueger, Tom Henighan, Rewon Child, Aditya Ramesh, Daniel Ziegler, Jeffrey Wu, Clemens Winter, Chris Hesse, Mark Chen, Eric Sigler, Mateusz Litwin, Scott Gray, Benjamin Chess, Jack Clark, Christopher Berner, Sam McCandlish, Alec Radford, Ilya Sutskever, and Dario Amodei. 2020.
\newblock \href {https://proceedings.neurips.cc/paper_files/paper/2020/file/1457c0d6bfcb4967418bfb8ac142f64a-Paper.pdf} {Language models are few-shot learners}.
\newblock In \emph{Advances in Neural Information Processing Systems}, volume~33, pages 1877--1901. Curran Associates, Inc.

\bibitem[{Dua et~al.(2022)Dua, Gupta, Singh, and Gardner}]{dua-etal-2022-successive}
Dheeru Dua, Shivanshu Gupta, Sameer Singh, and Matt Gardner. 2022.
\newblock \href {https://doi.org/10.18653/v1/2022.emnlp-main.81} {Successive prompting for decomposing complex questions}.
\newblock In \emph{Proceedings of the 2022 Conference on Empirical Methods in Natural Language Processing}, pages 1251--1265, Abu Dhabi, United Arab Emirates. Association for Computational Linguistics.

\bibitem[{Dubey et~al.(2024)Dubey, Jauhri, Pandey, Kadian, Al-Dahle, Letman, Mathur, Schelten, Yang, Fan, Goyal, Hartshorn, Yang, Mitra, Sravankumar, Korenev, Hinsvark, Rao, Zhang, Rodriguez, Gregerson, Spataru, Roziere, Biron, Tang, Chern, Caucheteux, Nayak, Bi, Marra, McConnell, Keller, Touret, Wu, Wong, Ferrer, Nikolaidis, Allonsius, Song, Pintz, Livshits, Esiobu, Choudhary, Mahajan, Garcia-Olano, Perino, Hupkes, Lakomkin, AlBadawy, Lobanova, Dinan, Smith, Radenovic, Zhang, Synnaeve, Lee, Anderson, Nail, Mialon, Pang, Cucurell, Nguyen, Korevaar, Xu, Touvron, Zarov, Ibarra, Kloumann, Misra, Evtimov, Copet, Lee, Geffert, Vranes, Park, Mahadeokar, Shah, van~der Linde, Billock, Hong, Lee, Fu, Chi, Huang, Liu, Wang, Yu, Bitton, Spisak, Park, Rocca, Johnstun, Saxe, Jia, Alwala, Upasani, Plawiak, Li, Heafield, Stone, El-Arini, Iyer, Malik, Chiu, Bhalla, Rantala-Yeary, van~der Maaten, Chen, Tan, Jenkins, Martin, Madaan, Malo, Blecher, Landzaat, de~Oliveira, Muzzi, Pasupuleti, Singh, Paluri, Kardas, Oldham, Rita,
  Pavlova, Kambadur, Lewis, Si, Singh, Hassan, Goyal, Torabi, Bashlykov, Bogoychev, Chatterji, Duchenne, Çelebi, Alrassy, Zhang, Li, Vasic, Weng, Bhargava, Dubal, Krishnan, Koura, Xu, He, Dong, Srinivasan, Ganapathy, Calderer, Cabral, Stojnic, Raileanu, Girdhar, Patel, Sauvestre, Polidoro, Sumbaly, Taylor, Silva, Hou, Wang, Hosseini, Chennabasappa, Singh, Bell, Kim, Edunov, Nie, Narang, Raparthy, Shen, Wan, Bhosale, Zhang, Vandenhende, Batra, Whitman, Sootla, Collot, Gururangan, Borodinsky, Herman, Fowler, Sheasha, Georgiou, Scialom, Speckbacher, Mihaylov, Xiao, Karn, Goswami, Gupta, Ramanathan, Kerkez, Gonguet, Do, Vogeti, Petrovic, Chu, Xiong, Fu, Meers, Martinet, Wang, Tan, Xie, Jia, Wang, Goldschlag, Gaur, Babaei, Wen, Song, Zhang, Li, Mao, Coudert, Yan, Chen, Papakipos, Singh, Grattafiori, Jain, Kelsey, Shajnfeld, Gangidi, Victoria, Goldstand, Menon, Sharma, Boesenberg, Vaughan, Baevski, Feinstein, Kallet, Sangani, Yunus, Lupu, Alvarado, Caples, Gu, Ho, Poulton, Ryan, Ramchandani, Franco, Saraf,
  Chowdhury, Gabriel, Bharambe, Eisenman, Yazdan, James, Maurer, Leonhardi, Huang, Loyd, Paola, Paranjape, Liu, Wu, Ni, Hancock, Wasti, Spence, Stojkovic, Gamido, Montalvo, Parker, Burton, Mejia, Wang, Kim, Zhou, Hu, Chu, Cai, Tindal, Feichtenhofer, Civin, Beaty, Kreymer, Li, Wyatt, Adkins, Xu, Testuggine, David, Parikh, Liskovich, Foss, Wang, Le, Holland, Dowling, Jamil, Montgomery, Presani, Hahn, Wood, Brinkman, Arcaute, Dunbar, Smothers, Sun, Kreuk, Tian, Ozgenel, Caggioni, Guzmán, Kanayet, Seide, Florez, Schwarz, Badeer, Swee, Halpern, Thattai, Herman, Sizov, Guangyi, Zhang, Lakshminarayanan, Shojanazeri, Zou, Wang, Zha, Habeeb, Rudolph, Suk, Aspegren, Goldman, Damlaj, Molybog, Tufanov, Veliche, Gat, Weissman, Geboski, Kohli, Asher, Gaya, Marcus, Tang, Chan, Zhen, Reizenstein, Teboul, Zhong, Jin, Yang, Cummings, Carvill, Shepard, McPhie, Torres, Ginsburg, Wang, Wu, U, Saxena, Prasad, Khandelwal, Zand, Matosich, Veeraraghavan, Michelena, Li, Huang, Chawla, Lakhotia, Huang, Chen, Garg, A, Silva, Bell,
  Zhang, Guo, Yu, Moshkovich, Wehrstedt, Khabsa, Avalani, Bhatt, Tsimpoukelli, Mankus, Hasson, Lennie, Reso, Groshev, Naumov, Lathi, Keneally, Seltzer, Valko, Restrepo, Patel, Vyatskov, Samvelyan, Clark, Macey, Wang, Hermoso, Metanat, Rastegari, Bansal, Santhanam, Parks, White, Bawa, Singhal, Egebo, Usunier, Laptev, Dong, Zhang, Cheng, Chernoguz, Hart, Salpekar, Kalinli, Kent, Parekh, Saab, Balaji, Rittner, Bontrager, Roux, Dollar, Zvyagina, Ratanchandani, Yuvraj, Liang, Alao, Rodriguez, Ayub, Murthy, Nayani, Mitra, Li, Hogan, Battey, Wang, Maheswari, Howes, Rinott, Bondu, Datta, Chugh, Hunt, Dhillon, Sidorov, Pan, Verma, Yamamoto, Ramaswamy, Lindsay, Lindsay, Feng, Lin, Zha, Shankar, Zhang, Zhang, Wang, Agarwal, Sajuyigbe, Chintala, Max, Chen, Kehoe, Satterfield, Govindaprasad, Gupta, Cho, Virk, Subramanian, Choudhury, Goldman, Remez, Glaser, Best, Kohler, Robinson, Li, Zhang, Matthews, Chou, Shaked, Vontimitta, Ajayi, Montanez, Mohan, Kumar, Mangla, Albiero, Ionescu, Poenaru, Mihailescu, Ivanov, Li, Wang,
  Jiang, Bouaziz, Constable, Tang, Wang, Wu, Wang, Xia, Wu, Gao, Chen, Hu, Jia, Qi, Li, Zhang, Zhang, Adi, Nam, Yu, Wang, Hao, Qian, He, Rait, DeVito, Rosnbrick, Wen, Yang, and Zhao}]{dubey2024llama3herdmodels}
Abhimanyu Dubey, Abhinav Jauhri, Abhinav Pandey, Abhishek Kadian, Ahmad Al-Dahle, Aiesha Letman, Akhil Mathur, Alan Schelten, Amy Yang, Angela Fan, Anirudh Goyal, Anthony Hartshorn, Aobo Yang, Archi Mitra, Archie Sravankumar, Artem Korenev, Arthur Hinsvark, Arun Rao, Aston Zhang, Aurelien Rodriguez, Austen Gregerson, Ava Spataru, Baptiste Roziere, Bethany Biron, Binh Tang, Bobbie Chern, Charlotte Caucheteux, Chaya Nayak, Chloe Bi, Chris Marra, Chris McConnell, Christian Keller, Christophe Touret, Chunyang Wu, Corinne Wong, Cristian~Canton Ferrer, Cyrus Nikolaidis, Damien Allonsius, Daniel Song, Danielle Pintz, Danny Livshits, David Esiobu, Dhruv Choudhary, Dhruv Mahajan, Diego Garcia-Olano, Diego Perino, Dieuwke Hupkes, Egor Lakomkin, Ehab AlBadawy, Elina Lobanova, Emily Dinan, Eric~Michael Smith, Filip Radenovic, Frank Zhang, Gabriel Synnaeve, Gabrielle Lee, Georgia~Lewis Anderson, Graeme Nail, Gregoire Mialon, Guan Pang, Guillem Cucurell, Hailey Nguyen, Hannah Korevaar, Hu~Xu, Hugo Touvron, Iliyan Zarov,
  Imanol~Arrieta Ibarra, Isabel Kloumann, Ishan Misra, Ivan Evtimov, Jade Copet, Jaewon Lee, Jan Geffert, Jana Vranes, Jason Park, Jay Mahadeokar, Jeet Shah, Jelmer van~der Linde, Jennifer Billock, Jenny Hong, Jenya Lee, Jeremy Fu, Jianfeng Chi, Jianyu Huang, Jiawen Liu, Jie Wang, Jiecao Yu, Joanna Bitton, Joe Spisak, Jongsoo Park, Joseph Rocca, Joshua Johnstun, Joshua Saxe, Junteng Jia, Kalyan~Vasuden Alwala, Kartikeya Upasani, Kate Plawiak, Ke~Li, Kenneth Heafield, Kevin Stone, Khalid El-Arini, Krithika Iyer, Kshitiz Malik, Kuenley Chiu, Kunal Bhalla, Lauren Rantala-Yeary, Laurens van~der Maaten, Lawrence Chen, Liang Tan, Liz Jenkins, Louis Martin, Lovish Madaan, Lubo Malo, Lukas Blecher, Lukas Landzaat, Luke de~Oliveira, Madeline Muzzi, Mahesh Pasupuleti, Mannat Singh, Manohar Paluri, Marcin Kardas, Mathew Oldham, Mathieu Rita, Maya Pavlova, Melanie Kambadur, Mike Lewis, Min Si, Mitesh~Kumar Singh, Mona Hassan, Naman Goyal, Narjes Torabi, Nikolay Bashlykov, Nikolay Bogoychev, Niladri Chatterji, Olivier
  Duchenne, Onur Çelebi, Patrick Alrassy, Pengchuan Zhang, Pengwei Li, Petar Vasic, Peter Weng, Prajjwal Bhargava, Pratik Dubal, Praveen Krishnan, Punit~Singh Koura, Puxin Xu, Qing He, Qingxiao Dong, Ragavan Srinivasan, Raj Ganapathy, Ramon Calderer, Ricardo~Silveira Cabral, Robert Stojnic, Roberta Raileanu, Rohit Girdhar, Rohit Patel, Romain Sauvestre, Ronnie Polidoro, Roshan Sumbaly, Ross Taylor, Ruan Silva, Rui Hou, Rui Wang, Saghar Hosseini, Sahana Chennabasappa, Sanjay Singh, Sean Bell, Seohyun~Sonia Kim, Sergey Edunov, Shaoliang Nie, Sharan Narang, Sharath Raparthy, Sheng Shen, Shengye Wan, Shruti Bhosale, Shun Zhang, Simon Vandenhende, Soumya Batra, Spencer Whitman, Sten Sootla, Stephane Collot, Suchin Gururangan, Sydney Borodinsky, Tamar Herman, Tara Fowler, Tarek Sheasha, Thomas Georgiou, Thomas Scialom, Tobias Speckbacher, Todor Mihaylov, Tong Xiao, Ujjwal Karn, Vedanuj Goswami, Vibhor Gupta, Vignesh Ramanathan, Viktor Kerkez, Vincent Gonguet, Virginie Do, Vish Vogeti, Vladan Petrovic, Weiwei Chu,
  Wenhan Xiong, Wenyin Fu, Whitney Meers, Xavier Martinet, Xiaodong Wang, Xiaoqing~Ellen Tan, Xinfeng Xie, Xuchao Jia, Xuewei Wang, Yaelle Goldschlag, Yashesh Gaur, Yasmine Babaei, Yi~Wen, Yiwen Song, Yuchen Zhang, Yue Li, Yuning Mao, Zacharie~Delpierre Coudert, Zheng Yan, Zhengxing Chen, Zoe Papakipos, Aaditya Singh, Aaron Grattafiori, Abha Jain, Adam Kelsey, Adam Shajnfeld, Adithya Gangidi, Adolfo Victoria, Ahuva Goldstand, Ajay Menon, Ajay Sharma, Alex Boesenberg, Alex Vaughan, Alexei Baevski, Allie Feinstein, Amanda Kallet, Amit Sangani, Anam Yunus, Andrei Lupu, Andres Alvarado, Andrew Caples, Andrew Gu, Andrew Ho, Andrew Poulton, Andrew Ryan, Ankit Ramchandani, Annie Franco, Aparajita Saraf, Arkabandhu Chowdhury, Ashley Gabriel, Ashwin Bharambe, Assaf Eisenman, Azadeh Yazdan, Beau James, Ben Maurer, Benjamin Leonhardi, Bernie Huang, Beth Loyd, Beto~De Paola, Bhargavi Paranjape, Bing Liu, Bo~Wu, Boyu Ni, Braden Hancock, Bram Wasti, Brandon Spence, Brani Stojkovic, Brian Gamido, Britt Montalvo, Carl
  Parker, Carly Burton, Catalina Mejia, Changhan Wang, Changkyu Kim, Chao Zhou, Chester Hu, Ching-Hsiang Chu, Chris Cai, Chris Tindal, Christoph Feichtenhofer, Damon Civin, Dana Beaty, Daniel Kreymer, Daniel Li, Danny Wyatt, David Adkins, David Xu, Davide Testuggine, Delia David, Devi Parikh, Diana Liskovich, Didem Foss, Dingkang Wang, Duc Le, Dustin Holland, Edward Dowling, Eissa Jamil, Elaine Montgomery, Eleonora Presani, Emily Hahn, Emily Wood, Erik Brinkman, Esteban Arcaute, Evan Dunbar, Evan Smothers, Fei Sun, Felix Kreuk, Feng Tian, Firat Ozgenel, Francesco Caggioni, Francisco Guzmán, Frank Kanayet, Frank Seide, Gabriela~Medina Florez, Gabriella Schwarz, Gada Badeer, Georgia Swee, Gil Halpern, Govind Thattai, Grant Herman, Grigory Sizov, Guangyi, Zhang, Guna Lakshminarayanan, Hamid Shojanazeri, Han Zou, Hannah Wang, Hanwen Zha, Haroun Habeeb, Harrison Rudolph, Helen Suk, Henry Aspegren, Hunter Goldman, Ibrahim Damlaj, Igor Molybog, Igor Tufanov, Irina-Elena Veliche, Itai Gat, Jake Weissman, James
  Geboski, James Kohli, Japhet Asher, Jean-Baptiste Gaya, Jeff Marcus, Jeff Tang, Jennifer Chan, Jenny Zhen, Jeremy Reizenstein, Jeremy Teboul, Jessica Zhong, Jian Jin, Jingyi Yang, Joe Cummings, Jon Carvill, Jon Shepard, Jonathan McPhie, Jonathan Torres, Josh Ginsburg, Junjie Wang, Kai Wu, Kam~Hou U, Karan Saxena, Karthik Prasad, Kartikay Khandelwal, Katayoun Zand, Kathy Matosich, Kaushik Veeraraghavan, Kelly Michelena, Keqian Li, Kun Huang, Kunal Chawla, Kushal Lakhotia, Kyle Huang, Lailin Chen, Lakshya Garg, Lavender A, Leandro Silva, Lee Bell, Lei Zhang, Liangpeng Guo, Licheng Yu, Liron Moshkovich, Luca Wehrstedt, Madian Khabsa, Manav Avalani, Manish Bhatt, Maria Tsimpoukelli, Martynas Mankus, Matan Hasson, Matthew Lennie, Matthias Reso, Maxim Groshev, Maxim Naumov, Maya Lathi, Meghan Keneally, Michael~L. Seltzer, Michal Valko, Michelle Restrepo, Mihir Patel, Mik Vyatskov, Mikayel Samvelyan, Mike Clark, Mike Macey, Mike Wang, Miquel~Jubert Hermoso, Mo~Metanat, Mohammad Rastegari, Munish Bansal, Nandhini
  Santhanam, Natascha Parks, Natasha White, Navyata Bawa, Nayan Singhal, Nick Egebo, Nicolas Usunier, Nikolay~Pavlovich Laptev, Ning Dong, Ning Zhang, Norman Cheng, Oleg Chernoguz, Olivia Hart, Omkar Salpekar, Ozlem Kalinli, Parkin Kent, Parth Parekh, Paul Saab, Pavan Balaji, Pedro Rittner, Philip Bontrager, Pierre Roux, Piotr Dollar, Polina Zvyagina, Prashant Ratanchandani, Pritish Yuvraj, Qian Liang, Rachad Alao, Rachel Rodriguez, Rafi Ayub, Raghotham Murthy, Raghu Nayani, Rahul Mitra, Raymond Li, Rebekkah Hogan, Robin Battey, Rocky Wang, Rohan Maheswari, Russ Howes, Ruty Rinott, Sai~Jayesh Bondu, Samyak Datta, Sara Chugh, Sara Hunt, Sargun Dhillon, Sasha Sidorov, Satadru Pan, Saurabh Verma, Seiji Yamamoto, Sharadh Ramaswamy, Shaun Lindsay, Shaun Lindsay, Sheng Feng, Shenghao Lin, Shengxin~Cindy Zha, Shiva Shankar, Shuqiang Zhang, Shuqiang Zhang, Sinong Wang, Sneha Agarwal, Soji Sajuyigbe, Soumith Chintala, Stephanie Max, Stephen Chen, Steve Kehoe, Steve Satterfield, Sudarshan Govindaprasad, Sumit Gupta,
  Sungmin Cho, Sunny Virk, Suraj Subramanian, Sy~Choudhury, Sydney Goldman, Tal Remez, Tamar Glaser, Tamara Best, Thilo Kohler, Thomas Robinson, Tianhe Li, Tianjun Zhang, Tim Matthews, Timothy Chou, Tzook Shaked, Varun Vontimitta, Victoria Ajayi, Victoria Montanez, Vijai Mohan, Vinay~Satish Kumar, Vishal Mangla, Vítor Albiero, Vlad Ionescu, Vlad Poenaru, Vlad~Tiberiu Mihailescu, Vladimir Ivanov, Wei Li, Wenchen Wang, Wenwen Jiang, Wes Bouaziz, Will Constable, Xiaocheng Tang, Xiaofang Wang, Xiaojian Wu, Xiaolan Wang, Xide Xia, Xilun Wu, Xinbo Gao, Yanjun Chen, Ye~Hu, Ye~Jia, Ye~Qi, Yenda Li, Yilin Zhang, Ying Zhang, Yossi Adi, Youngjin Nam, Yu, Wang, Yuchen Hao, Yundi Qian, Yuzi He, Zach Rait, Zachary DeVito, Zef Rosnbrick, Zhaoduo Wen, Zhenyu Yang, and Zhiwei Zhao. 2024.
\newblock \href {https://arxiv.org/abs/2407.21783} {The llama 3 herd of models}.
\newblock \emph{Preprint}, arXiv:2407.21783.

\bibitem[{Fu et~al.(2022)Fu, Peng, Sabharwal, Clark, and Khot}]{fu2022complexity}
Yao Fu, Hao Peng, Ashish Sabharwal, Peter Clark, and Tushar Khot. 2022.
\newblock Complexity-based prompting for multi-step reasoning.
\newblock In \emph{The Eleventh International Conference on Learning Representations}.

\bibitem[{Hendrycks et~al.(2021{\natexlab{a}})Hendrycks, Burns, Basart, Zou, Mazeika, Song, and Steinhardt}]{hendrycks2021measuring}
Dan Hendrycks, Collin Burns, Steven Basart, Andy Zou, Mantas Mazeika, Dawn Song, and Jacob Steinhardt. 2021{\natexlab{a}}.
\newblock \href {https://openreview.net/forum?id=d7KBjmI3GmQ} {Measuring massive multitask language understanding}.
\newblock In \emph{International Conference on Learning Representations}.

\bibitem[{Hendrycks et~al.(2021{\natexlab{b}})Hendrycks, Burns, Kadavath, Arora, Basart, Tang, Song, and Steinhardt}]{hendrycksmath2021}
Dan Hendrycks, Collin Burns, Saurav Kadavath, Akul Arora, Steven Basart, Eric Tang, Dawn Song, and Jacob Steinhardt. 2021{\natexlab{b}}.
\newblock Measuring mathematical problem solving with the math dataset.
\newblock \emph{NeurIPS}.

\bibitem[{Huang et~al.(2024)Huang, Chen, Mishra, Zheng, Yu, Song, and Zhou}]{huang2024large}
Jie Huang, Xinyun Chen, Swaroop Mishra, Huaixiu~Steven Zheng, Adams~Wei Yu, Xinying Song, and Denny Zhou. 2024.
\newblock \href {https://openreview.net/forum?id=IkmD3fKBPQ} {Large language models cannot self-correct reasoning yet}.
\newblock In \emph{The Twelfth International Conference on Learning Representations}.

\bibitem[{Khot et~al.(2023)Khot, Trivedi, Finlayson, Fu, Richardson, Clark, and Sabharwal}]{khot2023decomposed}
Tushar Khot, Harsh Trivedi, Matthew Finlayson, Yao Fu, Kyle Richardson, Peter Clark, and Ashish Sabharwal. 2023.
\newblock \href {https://openreview.net/forum?id=_nGgzQjzaRy} {Decomposed prompting: A modular approach for solving complex tasks}.
\newblock In \emph{The Eleventh International Conference on Learning Representations}.

\bibitem[{Kojima et~al.(2022)Kojima, Gu, Reid, Matsuo, and Iwasawa}]{kojima2022large}
Takeshi Kojima, Shixiang~Shane Gu, Machel Reid, Yutaka Matsuo, and Yusuke Iwasawa. 2022.
\newblock \href {https://openreview.net/forum?id=e2TBb5y0yFf} {Large language models are zero-shot reasoners}.
\newblock In \emph{Advances in Neural Information Processing Systems}.

\bibitem[{Madaan et~al.(2023)Madaan, Tandon, Gupta, Hallinan, Gao, Wiegreffe, Alon, Dziri, Prabhumoye, Yang, Gupta, Majumder, Hermann, Welleck, Yazdanbakhsh, and Clark}]{madaan2023selfrefine}
Aman Madaan, Niket Tandon, Prakhar Gupta, Skyler Hallinan, Luyu Gao, Sarah Wiegreffe, Uri Alon, Nouha Dziri, Shrimai Prabhumoye, Yiming Yang, Shashank Gupta, Bodhisattwa~Prasad Majumder, Katherine Hermann, Sean Welleck, Amir Yazdanbakhsh, and Peter Clark. 2023.
\newblock \href {https://openreview.net/forum?id=S37hOerQLB} {Self-refine: Iterative refinement with self-feedback}.
\newblock In \emph{Thirty-seventh Conference on Neural Information Processing Systems}.

\bibitem[{\mbox{Gemma Team} et~al.(2024)\mbox{Gemma Team}, Mesnard, Hardin, Dadashi, Bhupatiraju, Pathak, Sifre, Rivière, Kale, Love, Tafti, Hussenot, Sessa, Chowdhery, Roberts, Barua, Botev, Castro-Ros, Slone, Héliou, Tacchetti, Bulanova, Paterson, Tsai, Shahriari, Lan, Choquette-Choo, Crepy, Cer, Ippolito, Reid, Buchatskaya, Ni, Noland, Yan, Tucker, Muraru, Rozhdestvenskiy, Michalewski, Tenney, Grishchenko, Austin, Keeling, Labanowski, Lespiau, Stanway, Brennan, Chen, Ferret, Chiu, Mao-Jones, Lee, Yu, Millican, Sjoesund, Lee, Dixon, Reid, Mikuła, Wirth, Sharman, Chinaev, Thain, Bachem, Chang, Wahltinez, Bailey, Michel, Yotov, Chaabouni, Comanescu, Jana, Anil, McIlroy, Liu, Mullins, Smith, Borgeaud, Girgin, Douglas, Pandya, Shakeri, De, Klimenko, Hennigan, Feinberg, Stokowiec, hui Chen, Ahmed, Gong, Warkentin, Peran, Giang, Farabet, Vinyals, Dean, Kavukcuoglu, Hassabis, Ghahramani, Eck, Barral, Pereira, Collins, Joulin, Fiedel, Senter, Andreev, and Kenealy}]{gemmateam2024gemma}
\mbox{Gemma Team}, Thomas Mesnard, Cassidy Hardin, Robert Dadashi, Surya Bhupatiraju, Shreya Pathak, Laurent Sifre, Morgane Rivière, Mihir~Sanjay Kale, Juliette Love, Pouya Tafti, Léonard Hussenot, Pier~Giuseppe Sessa, Aakanksha Chowdhery, Adam Roberts, Aditya Barua, Alex Botev, Alex Castro-Ros, Ambrose Slone, Amélie Héliou, Andrea Tacchetti, Anna Bulanova, Antonia Paterson, Beth Tsai, Bobak Shahriari, Charline~Le Lan, Christopher~A. Choquette-Choo, Clément Crepy, Daniel Cer, Daphne Ippolito, David Reid, Elena Buchatskaya, Eric Ni, Eric Noland, Geng Yan, George Tucker, George-Christian Muraru, Grigory Rozhdestvenskiy, Henryk Michalewski, Ian Tenney, Ivan Grishchenko, Jacob Austin, James Keeling, Jane Labanowski, Jean-Baptiste Lespiau, Jeff Stanway, Jenny Brennan, Jeremy Chen, Johan Ferret, Justin Chiu, Justin Mao-Jones, Katherine Lee, Kathy Yu, Katie Millican, Lars~Lowe Sjoesund, Lisa Lee, Lucas Dixon, Machel Reid, Maciej Mikuła, Mateo Wirth, Michael Sharman, Nikolai Chinaev, Nithum Thain, Olivier
  Bachem, Oscar Chang, Oscar Wahltinez, Paige Bailey, Paul Michel, Petko Yotov, Rahma Chaabouni, Ramona Comanescu, Reena Jana, Rohan Anil, Ross McIlroy, Ruibo Liu, Ryan Mullins, Samuel~L Smith, Sebastian Borgeaud, Sertan Girgin, Sholto Douglas, Shree Pandya, Siamak Shakeri, Soham De, Ted Klimenko, Tom Hennigan, Vlad Feinberg, Wojciech Stokowiec, Yu~hui Chen, Zafarali Ahmed, Zhitao Gong, Tris Warkentin, Ludovic Peran, Minh Giang, Clément Farabet, Oriol Vinyals, Jeff Dean, Koray Kavukcuoglu, Demis Hassabis, Zoubin Ghahramani, Douglas Eck, Joelle Barral, Fernando Pereira, Eli Collins, Armand Joulin, Noah Fiedel, Evan Senter, Alek Andreev, and Kathleen Kenealy. 2024.
\newblock \href {https://arxiv.org/abs/2403.08295} {Gemma: Open models based on gemini research and technology}.
\newblock \emph{Preprint}, arXiv:2403.08295.

\bibitem[{Srivastava et~al.(2023)Srivastava, Rastogi, Rao, Shoeb, Abid, Fisch, Brown, Santoro, Gupta, Garriga-Alonso, Kluska, Lewkowycz, Agarwal, Power, Ray, Warstadt, Kocurek, Safaya, Tazarv, Xiang, Parrish, Nie, Hussain, Askell, Dsouza, Slone, Rahane, Iyer, Andreassen, Madotto, Santilli, Stuhlm{\"u}ller, Dai, La, Lampinen, Zou, Jiang, Chen, Vuong, Gupta, Gottardi, Norelli, Venkatesh, Gholamidavoodi, Tabassum, Menezes, Kirubarajan, Mullokandov, Sabharwal, Herrick, Efrat, Erdem, Karaka{\c{s}}, Roberts, Loe, Zoph, Bojanowski, {\"O}zyurt, Hedayatnia, Neyshabur, Inden, Stein, Ekmekci, Lin, Howald, Orinion, Diao, Dour, Stinson, Argueta, Ferri, Singh, Rathkopf, Meng, Baral, Wu, Callison-Burch, Waites, Voigt, Manning, Potts, Ramirez, Rivera, Siro, Raffel, Ashcraft, Garbacea, Sileo, Garrette, Hendrycks, Kilman, Roth, Freeman, Khashabi, Levy, Gonz{\'a}lez, Perszyk, Hernandez, Chen, Ippolito, Gilboa, Dohan, Drakard, Jurgens, Datta, Ganguli, Emelin, Kleyko, Yuret, Chen, Tam, Hupkes, Misra, Buzan, Mollo, Yang, Lee,
  Schrader, Shutova, Cubuk, Segal, Hagerman, Barnes, Donoway, Pavlick, Rodol{\`a}, Lam, Chu, Tang, Erdem, Chang, Chi, Dyer, Jerzak, Kim, Manyasi, Zheltonozhskii, Xia, Siar, Mart{\'\i}nez-Plumed, Happ{\'e}, Chollet, Rong, Mishra, Winata, de~Melo, Kruszewski, Parascandolo, Mariani, Wang, Jaimovitch-Lopez, Betz, Gur-Ari, Galijasevic, Kim, Rashkin, Hajishirzi, Mehta, Bogar, Shevlin, Schuetze, Yakura, Zhang, Wong, Ng, Noble, Jumelet, Geissinger, Kernion, Hilton, Lee, Fisac, Simon, Koppel, Zheng, Zou, Kocon, Thompson, Wingfield, Kaplan, Radom, Sohl-Dickstein, Phang, Wei, Yosinski, Novikova, Bosscher, Marsh, Kim, Taal, Engel, Alabi, Xu, Song, Tang, Waweru, Burden, Miller, Balis, Batchelder, Berant, Frohberg, Rozen, Hernandez-Orallo, Boudeman, Guerr, Jones, Tenenbaum, Rule, Chua, Kanclerz, Livescu, Krauth, Gopalakrishnan, Ignatyeva, Markert, Dhole, Gimpel, Omondi, Mathewson, Chiafullo, Shkaruta, Shridhar, McDonell, Richardson, Reynolds, Gao, Zhang, Dugan, Qin, Contreras-Ochando, Morency, Moschella, Lam, Noble,
  Schmidt, He, Oliveros-Col{\'o}n, Metz, Senel, Bosma, Sap, Hoeve, Farooqi, Faruqui, Mazeika, Baturan, Marelli, Maru, Ramirez-Quintana, Tolkiehn, Giulianelli, Lewis, Potthast, Leavitt, Hagen, Schubert, Baitemirova, Arnaud, McElrath, Yee, Cohen, Gu, Ivanitskiy, Starritt, Strube, Sw{\k{e}}drowski, Bevilacqua, Yasunaga, Kale, Cain, Xu, Suzgun, Walker, Tiwari, Bansal, Aminnaseri, Geva, Gheini, T, Peng, Chi, Lee, Krakover, Cameron, Roberts, Doiron, Martinez, Nangia, Deckers, Muennighoff, Keskar, Iyer, Constant, Fiedel, Wen, Zhang, Agha, Elbaghdadi, Levy, Evans, Casares, Doshi, Fung, Liang, Vicol, Alipoormolabashi, Liao, Liang, Chang, Eckersley, Htut, Hwang, Mi{\l}kowski, Patil, Pezeshkpour, Oli, Mei, Lyu, Chen, Banjade, Rudolph, Gabriel, Habacker, Risco, Milli{\`e}re, Garg, Barnes, Saurous, Arakawa, Raymaekers, Frank, Sikand, Novak, Sitelew, Bras, Liu, Jacobs, Zhang, Salakhutdinov, Chi, Lee, Stovall, Teehan, Yang, Singh, Mohammad, Anand, Dillavou, Shleifer, Wiseman, Gruetter, Bowman, Schoenholz, Han, Kwatra, Rous,
  Ghazarian, Ghosh, Casey, Bischoff, Gehrmann, Schuster, Sadeghi, Hamdan, Zhou, Srivastava, Shi, Singh, Asaadi, Gu, Pachchigar, Toshniwal, Upadhyay, Debnath, Shakeri, Thormeyer, Melzi, Reddy, Makini, Lee, Torene, Hatwar, Dehaene, Divic, Ermon, Biderman, Lin, Prasad, Piantadosi, Shieber, Misherghi, Kiritchenko, Mishra, Linzen, Schuster, Li, Yu, Ali, Hashimoto, Wu, Desbordes, Rothschild, Phan, Wang, Nkinyili, Schick, Kornev, Tunduny, Gerstenberg, Chang, Neeraj, Khot, Shultz, Shaham, Misra, Demberg, Nyamai, Raunak, Ramasesh, vinay~uday prabhu, Padmakumar, Srikumar, Fedus, Saunders, Zhang, Vossen, Ren, Tong, Zhao, Wu, Shen, Yaghoobzadeh, Lakretz, Song, Bahri, Choi, Yang, Hao, Chen, Belinkov, Hou, Hou, Bai, Seid, Zhao, Wang, Wang, Wang, and Wu}]{srivastava2023beyond}
Aarohi Srivastava, Abhinav Rastogi, Abhishek Rao, Abu Awal~Md Shoeb, Abubakar Abid, Adam Fisch, Adam~R. Brown, Adam Santoro, Aditya Gupta, Adri{\`a} Garriga-Alonso, Agnieszka Kluska, Aitor Lewkowycz, Akshat Agarwal, Alethea Power, Alex Ray, Alex Warstadt, Alexander~W. Kocurek, Ali Safaya, Ali Tazarv, Alice Xiang, Alicia Parrish, Allen Nie, Aman Hussain, Amanda Askell, Amanda Dsouza, Ambrose Slone, Ameet Rahane, Anantharaman~S. Iyer, Anders~Johan Andreassen, Andrea Madotto, Andrea Santilli, Andreas Stuhlm{\"u}ller, Andrew~M. Dai, Andrew La, Andrew Lampinen, Andy Zou, Angela Jiang, Angelica Chen, Anh Vuong, Animesh Gupta, Anna Gottardi, Antonio Norelli, Anu Venkatesh, Arash Gholamidavoodi, Arfa Tabassum, Arul Menezes, Arun Kirubarajan, Asher Mullokandov, Ashish Sabharwal, Austin Herrick, Avia Efrat, Aykut Erdem, Ayla Karaka{\c{s}}, B.~Ryan Roberts, Bao~Sheng Loe, Barret Zoph, Bart{\l}omiej Bojanowski, Batuhan {\"O}zyurt, Behnam Hedayatnia, Behnam Neyshabur, Benjamin Inden, Benno Stein, Berk Ekmekci, Bill~Yuchen
  Lin, Blake Howald, Bryan Orinion, Cameron Diao, Cameron Dour, Catherine Stinson, Cedrick Argueta, Cesar Ferri, Chandan Singh, Charles Rathkopf, Chenlin Meng, Chitta Baral, Chiyu Wu, Chris Callison-Burch, Christopher Waites, Christian Voigt, Christopher~D Manning, Christopher Potts, Cindy Ramirez, Clara~E. Rivera, Clemencia Siro, Colin Raffel, Courtney Ashcraft, Cristina Garbacea, Damien Sileo, Dan Garrette, Dan Hendrycks, Dan Kilman, Dan Roth, C.~Daniel Freeman, Daniel Khashabi, Daniel Levy, Daniel~Mosegu{\'\i} Gonz{\'a}lez, Danielle Perszyk, Danny Hernandez, Danqi Chen, Daphne Ippolito, Dar Gilboa, David Dohan, David Drakard, David Jurgens, Debajyoti Datta, Deep Ganguli, Denis Emelin, Denis Kleyko, Deniz Yuret, Derek Chen, Derek Tam, Dieuwke Hupkes, Diganta Misra, Dilyar Buzan, Dimitri~Coelho Mollo, Diyi Yang, Dong-Ho Lee, Dylan Schrader, Ekaterina Shutova, Ekin~Dogus Cubuk, Elad Segal, Eleanor Hagerman, Elizabeth Barnes, Elizabeth Donoway, Ellie Pavlick, Emanuele Rodol{\`a}, Emma Lam, Eric Chu, Eric Tang,
  Erkut Erdem, Ernie Chang, Ethan~A Chi, Ethan Dyer, Ethan Jerzak, Ethan Kim, Eunice~Engefu Manyasi, Evgenii Zheltonozhskii, Fanyue Xia, Fatemeh Siar, Fernando Mart{\'\i}nez-Plumed, Francesca Happ{\'e}, Francois Chollet, Frieda Rong, Gaurav Mishra, Genta~Indra Winata, Gerard de~Melo, Germ{\'a}n Kruszewski, Giambattista Parascandolo, Giorgio Mariani, Gloria~Xinyue Wang, Gonzalo Jaimovitch-Lopez, Gregor Betz, Guy Gur-Ari, Hana Galijasevic, Hannah Kim, Hannah Rashkin, Hannaneh Hajishirzi, Harsh Mehta, Hayden Bogar, Henry Francis~Anthony Shevlin, Hinrich Schuetze, Hiromu Yakura, Hongming Zhang, Hugh~Mee Wong, Ian Ng, Isaac Noble, Jaap Jumelet, Jack Geissinger, Jackson Kernion, Jacob Hilton, Jaehoon Lee, Jaime~Fern{\'a}ndez Fisac, James~B Simon, James Koppel, James Zheng, James Zou, Jan Kocon, Jana Thompson, Janelle Wingfield, Jared Kaplan, Jarema Radom, Jascha Sohl-Dickstein, Jason Phang, Jason Wei, Jason Yosinski, Jekaterina Novikova, Jelle Bosscher, Jennifer Marsh, Jeremy Kim, Jeroen Taal, Jesse Engel, Jesujoba
  Alabi, Jiacheng Xu, Jiaming Song, Jillian Tang, Joan Waweru, John Burden, John Miller, John~U. Balis, Jonathan Batchelder, Jonathan Berant, J{\"o}rg Frohberg, Jos Rozen, Jose Hernandez-Orallo, Joseph Boudeman, Joseph Guerr, Joseph Jones, Joshua~B. Tenenbaum, Joshua~S. Rule, Joyce Chua, Kamil Kanclerz, Karen Livescu, Karl Krauth, Karthik Gopalakrishnan, Katerina Ignatyeva, Katja Markert, Kaustubh Dhole, Kevin Gimpel, Kevin Omondi, Kory~Wallace Mathewson, Kristen Chiafullo, Ksenia Shkaruta, Kumar Shridhar, Kyle McDonell, Kyle Richardson, Laria Reynolds, Leo Gao, Li~Zhang, Liam Dugan, Lianhui Qin, Lidia Contreras-Ochando, Louis-Philippe Morency, Luca Moschella, Lucas Lam, Lucy Noble, Ludwig Schmidt, Luheng He, Luis Oliveros-Col{\'o}n, Luke Metz, L{\"u}tfi~Kerem Senel, Maarten Bosma, Maarten Sap, Maartje~Ter Hoeve, Maheen Farooqi, Manaal Faruqui, Mantas Mazeika, Marco Baturan, Marco Marelli, Marco Maru, Maria~Jose Ramirez-Quintana, Marie Tolkiehn, Mario Giulianelli, Martha Lewis, Martin Potthast, Matthew~L
  Leavitt, Matthias Hagen, M{\'a}ty{\'a}s Schubert, Medina~Orduna Baitemirova, Melody Arnaud, Melvin McElrath, Michael~Andrew Yee, Michael Cohen, Michael Gu, Michael Ivanitskiy, Michael Starritt, Michael Strube, Micha{\l} Sw{\k{e}}drowski, Michele Bevilacqua, Michihiro Yasunaga, Mihir Kale, Mike Cain, Mimee Xu, Mirac Suzgun, Mitch Walker, Mo~Tiwari, Mohit Bansal, Moin Aminnaseri, Mor Geva, Mozhdeh Gheini, Mukund~Varma T, Nanyun Peng, Nathan~Andrew Chi, Nayeon Lee, Neta Gur-Ari Krakover, Nicholas Cameron, Nicholas Roberts, Nick Doiron, Nicole Martinez, Nikita Nangia, Niklas Deckers, Niklas Muennighoff, Nitish~Shirish Keskar, Niveditha~S. Iyer, Noah Constant, Noah Fiedel, Nuan Wen, Oliver Zhang, Omar Agha, Omar Elbaghdadi, Omer Levy, Owain Evans, Pablo Antonio~Moreno Casares, Parth Doshi, Pascale Fung, Paul~Pu Liang, Paul Vicol, Pegah Alipoormolabashi, Peiyuan Liao, Percy Liang, Peter~W Chang, Peter Eckersley, Phu~Mon Htut, Pinyu Hwang, Piotr Mi{\l}kowski, Piyush Patil, Pouya Pezeshkpour, Priti Oli, Qiaozhu
  Mei, Qing Lyu, Qinlang Chen, Rabin Banjade, Rachel~Etta Rudolph, Raefer Gabriel, Rahel Habacker, Ramon Risco, Rapha{\"e}l Milli{\`e}re, Rhythm Garg, Richard Barnes, Rif~A. Saurous, Riku Arakawa, Robbe Raymaekers, Robert Frank, Rohan Sikand, Roman Novak, Roman Sitelew, Ronan~Le Bras, Rosanne Liu, Rowan Jacobs, Rui Zhang, Russ Salakhutdinov, Ryan~Andrew Chi, Seungjae~Ryan Lee, Ryan Stovall, Ryan Teehan, Rylan Yang, Sahib Singh, Saif~M. Mohammad, Sajant Anand, Sam Dillavou, Sam Shleifer, Sam Wiseman, Samuel Gruetter, Samuel~R. Bowman, Samuel~Stern Schoenholz, Sanghyun Han, Sanjeev Kwatra, Sarah~A. Rous, Sarik Ghazarian, Sayan Ghosh, Sean Casey, Sebastian Bischoff, Sebastian Gehrmann, Sebastian Schuster, Sepideh Sadeghi, Shadi Hamdan, Sharon Zhou, Shashank Srivastava, Sherry Shi, Shikhar Singh, Shima Asaadi, Shixiang~Shane Gu, Shubh Pachchigar, Shubham Toshniwal, Shyam Upadhyay, Shyamolima~Shammie Debnath, Siamak Shakeri, Simon Thormeyer, Simone Melzi, Siva Reddy, Sneha~Priscilla Makini, Soo-Hwan Lee, Spencer
  Torene, Sriharsha Hatwar, Stanislas Dehaene, Stefan Divic, Stefano Ermon, Stella Biderman, Stephanie Lin, Stephen Prasad, Steven Piantadosi, Stuart Shieber, Summer Misherghi, Svetlana Kiritchenko, Swaroop Mishra, Tal Linzen, Tal Schuster, Tao Li, Tao Yu, Tariq Ali, Tatsunori Hashimoto, Te-Lin Wu, Th{\'e}o Desbordes, Theodore Rothschild, Thomas Phan, Tianle Wang, Tiberius Nkinyili, Timo Schick, Timofei Kornev, Titus Tunduny, Tobias Gerstenberg, Trenton Chang, Trishala Neeraj, Tushar Khot, Tyler Shultz, Uri Shaham, Vedant Misra, Vera Demberg, Victoria Nyamai, Vikas Raunak, Vinay~Venkatesh Ramasesh, vinay~uday prabhu, Vishakh Padmakumar, Vivek Srikumar, William Fedus, William Saunders, William Zhang, Wout Vossen, Xiang Ren, Xiaoyu Tong, Xinran Zhao, Xinyi Wu, Xudong Shen, Yadollah Yaghoobzadeh, Yair Lakretz, Yangqiu Song, Yasaman Bahri, Yejin Choi, Yichi Yang, Yiding Hao, Yifu Chen, Yonatan Belinkov, Yu~Hou, Yufang Hou, Yuntao Bai, Zachary Seid, Zhuoye Zhao, Zijian Wang, Zijie~J. Wang, Zirui Wang, and Ziyi Wu.
  2023.
\newblock \href {https://openreview.net/forum?id=uyTL5Bvosj} {Beyond the imitation game: Quantifying and extrapolating the capabilities of language models}.
\newblock \emph{Transactions on Machine Learning Research}.

\bibitem[{Suzgun et~al.(2023)Suzgun, Scales, Sch{\"a}rli, Gehrmann, Tay, Chung, Chowdhery, Le, Chi, Zhou, and Wei}]{suzgun-etal-2023-challenging}
Mirac Suzgun, Nathan Scales, Nathanael Sch{\"a}rli, Sebastian Gehrmann, Yi~Tay, Hyung~Won Chung, Aakanksha Chowdhery, Quoc Le, Ed~Chi, Denny Zhou, and Jason Wei. 2023.
\newblock \href {https://doi.org/10.18653/v1/2023.findings-acl.824} {Challenging {BIG}-bench tasks and whether chain-of-thought can solve them}.
\newblock In \emph{Findings of the Association for Computational Linguistics: ACL 2023}, pages 13003--13051, Toronto, Canada. Association for Computational Linguistics.

\bibitem[{Touvron et~al.(2023)Touvron, Martin, Stone, Albert, Almahairi, Babaei, Bashlykov, Batra, Bhargava, Bhosale, Bikel, Blecher, Ferrer, Chen, Cucurull, Esiobu, Fernandes, Fu, Fu, Fuller, Gao, Goswami, Goyal, Hartshorn, Hosseini, Hou, Inan, Kardas, Kerkez, Khabsa, Kloumann, Korenev, Koura, Lachaux, Lavril, Lee, Liskovich, Lu, Mao, Martinet, Mihaylov, Mishra, Molybog, Nie, Poulton, Reizenstein, Rungta, Saladi, Schelten, Silva, Smith, Subramanian, Tan, Tang, Taylor, Williams, Kuan, Xu, Yan, Zarov, Zhang, Fan, Kambadur, Narang, Rodriguez, Stojnic, Edunov, and Scialom}]{touvron2023llama}
Hugo Touvron, Louis Martin, Kevin Stone, Peter Albert, Amjad Almahairi, Yasmine Babaei, Nikolay Bashlykov, Soumya Batra, Prajjwal Bhargava, Shruti Bhosale, Dan Bikel, Lukas Blecher, Cristian~Canton Ferrer, Moya Chen, Guillem Cucurull, David Esiobu, Jude Fernandes, Jeremy Fu, Wenyin Fu, Brian Fuller, Cynthia Gao, Vedanuj Goswami, Naman Goyal, Anthony Hartshorn, Saghar Hosseini, Rui Hou, Hakan Inan, Marcin Kardas, Viktor Kerkez, Madian Khabsa, Isabel Kloumann, Artem Korenev, Punit~Singh Koura, Marie-Anne Lachaux, Thibaut Lavril, Jenya Lee, Diana Liskovich, Yinghai Lu, Yuning Mao, Xavier Martinet, Todor Mihaylov, Pushkar Mishra, Igor Molybog, Yixin Nie, Andrew Poulton, Jeremy Reizenstein, Rashi Rungta, Kalyan Saladi, Alan Schelten, Ruan Silva, Eric~Michael Smith, Ranjan Subramanian, Xiaoqing~Ellen Tan, Binh Tang, Ross Taylor, Adina Williams, Jian~Xiang Kuan, Puxin Xu, Zheng Yan, Iliyan Zarov, Yuchen Zhang, Angela Fan, Melanie Kambadur, Sharan Narang, Aurelien Rodriguez, Robert Stojnic, Sergey Edunov, and Thomas
  Scialom. 2023.
\newblock \href {https://arxiv.org/abs/2307.09288} {Llama 2: Open foundation and fine-tuned chat models}.
\newblock \emph{Preprint}, arXiv:2307.09288.

\bibitem[{Wang et~al.(2023{\natexlab{a}})Wang, Xu, Lan, Hu, Lan, Lee, and Lim}]{wang-etal-2023-plan}
Lei Wang, Wanyu Xu, Yihuai Lan, Zhiqiang Hu, Yunshi Lan, Roy Ka-Wei Lee, and Ee-Peng Lim. 2023{\natexlab{a}}.
\newblock \href {https://doi.org/10.18653/v1/2023.acl-long.147} {Plan-and-solve prompting: Improving zero-shot chain-of-thought reasoning by large language models}.
\newblock In \emph{Proceedings of the 61st Annual Meeting of the Association for Computational Linguistics (Volume 1: Long Papers)}, pages 2609--2634, Toronto, Canada. Association for Computational Linguistics.

\bibitem[{Wang et~al.(2023{\natexlab{b}})Wang, Wei, Schuurmans, Le, Chi, Narang, Chowdhery, and Zhou}]{wang2023selfconsistency}
Xuezhi Wang, Jason Wei, Dale Schuurmans, Quoc~V Le, Ed~H. Chi, Sharan Narang, Aakanksha Chowdhery, and Denny Zhou. 2023{\natexlab{b}}.
\newblock \href {https://openreview.net/forum?id=1PL1NIMMrw} {Self-consistency improves chain of thought reasoning in language models}.
\newblock In \emph{The Eleventh International Conference on Learning Representations}.

\bibitem[{Wei et~al.(2022)Wei, Wang, Schuurmans, Bosma, ichter, Xia, Chi, Le, and Zhou}]{NEURIPS2022_9d560961}
Jason Wei, Xuezhi Wang, Dale Schuurmans, Maarten Bosma, brian ichter, Fei Xia, Ed~Chi, Quoc~V Le, and Denny Zhou. 2022.
\newblock \href {https://proceedings.neurips.cc/paper_files/paper/2022/file/9d5609613524ecf4f15af0f7b31abca4-Paper-Conference.pdf} {Chain-of-thought prompting elicits reasoning in large language models}.
\newblock In \emph{Advances in Neural Information Processing Systems}, volume~35, pages 24824--24837. Curran Associates, Inc.

\bibitem[{Yao et~al.(2023)Yao, Yu, Zhao, Shafran, Griffiths, Cao, and Narasimhan}]{yao2023tree}
Shunyu Yao, Dian Yu, Jeffrey Zhao, Izhak Shafran, Thomas~L. Griffiths, Yuan Cao, and Karthik~R Narasimhan. 2023.
\newblock \href {https://openreview.net/forum?id=5Xc1ecxO1h} {Tree of thoughts: Deliberate problem solving with large language models}.
\newblock In \emph{Thirty-seventh Conference on Neural Information Processing Systems}.

\bibitem[{Zhang et~al.(2023)Zhang, Zhang, Li, and Smola}]{zhang2023automatic}
Zhuosheng Zhang, Aston Zhang, Mu~Li, and Alex Smola. 2023.
\newblock \href {https://openreview.net/forum?id=5NTt8GFjUHkr} {Automatic chain of thought prompting in large language models}.
\newblock In \emph{The Eleventh International Conference on Learning Representations}.

\bibitem[{Zhou et~al.(2023)Zhou, Sch{\"a}rli, Hou, Wei, Scales, Wang, Schuurmans, Cui, Bousquet, Le, and Chi}]{zhou2023leasttomost}
Denny Zhou, Nathanael Sch{\"a}rli, Le~Hou, Jason Wei, Nathan Scales, Xuezhi Wang, Dale Schuurmans, Claire Cui, Olivier Bousquet, Quoc~V Le, and Ed~H. Chi. 2023.
\newblock \href {https://openreview.net/forum?id=WZH7099tgfM} {Least-to-most prompting enables complex reasoning in large language models}.
\newblock In \emph{The Eleventh International Conference on Learning Representations}.

\end{thebibliography}
% Custom bibliography entries only
% \bibliography{custom}

\appendix

\section{Clarifications}\label{app:clarifications}

\subsection{Canonical Tasks}\label{app:top2-2}
\begin{figure*}[!ht]
  \includegraphics[width=\textwidth]{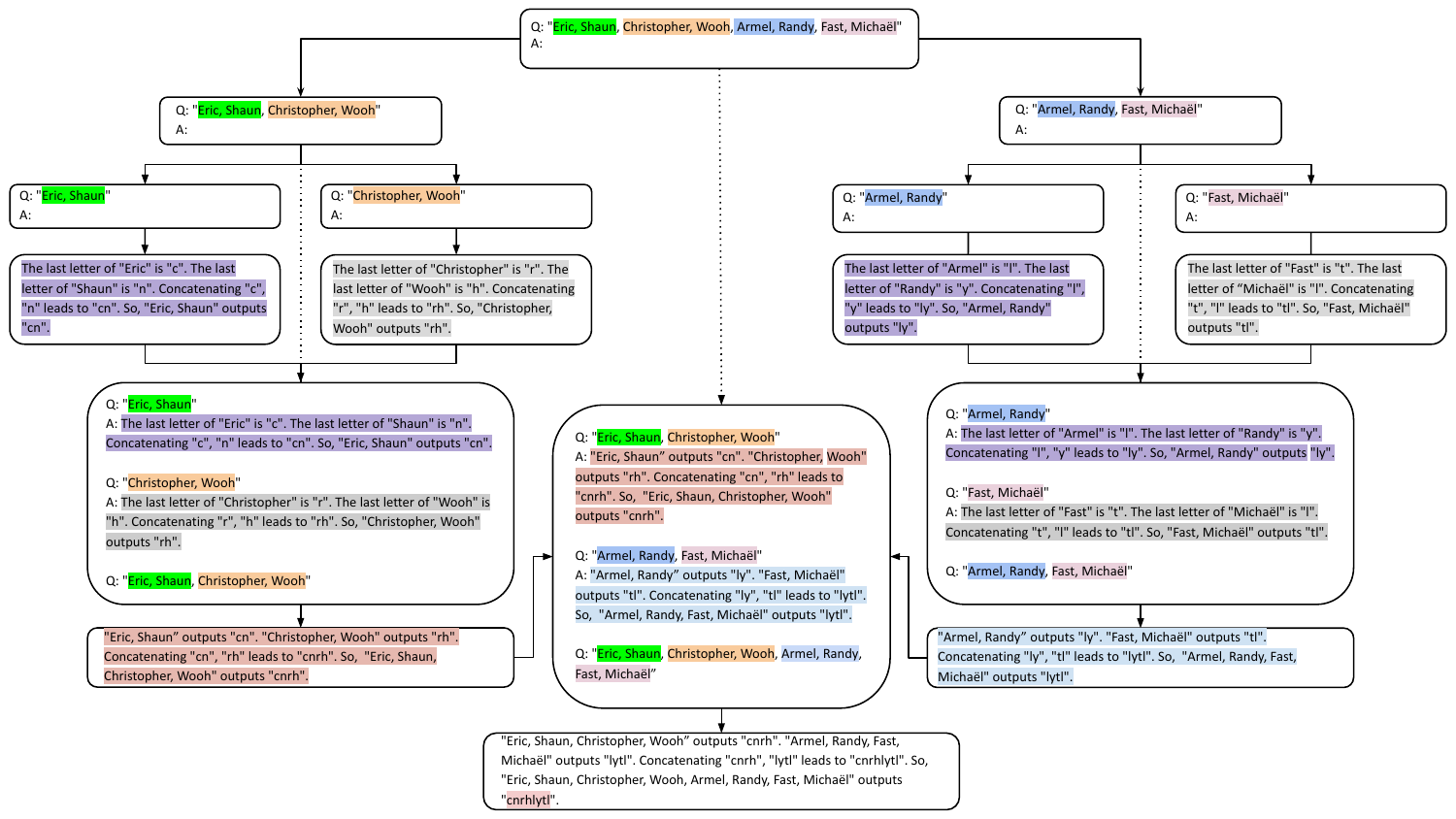}
  \caption{Overview of ToP (2, 2) for Last Letter Concatenation. The list of words is divided into 2 sublists which are recursively divided into two sublists. The problems at the leaves, which consist into concatenating the last letters of 2-word lists are solved first. The solutions are then merged in a bottom-up way until the main instance is solved.}
  \label{fig:deeper}
\end{figure*}

In Figure~\ref{fig:fig1} we showed how to apply $ToP~(2, 1)$ to an instance of Last Letter Concatenation. We illustrate how $ToP~(2, 2)$ would look for concatenating the last letters of a list of 8 words in Figure~\ref{fig:deeper}. The decomposition is done on two levels, the leaves being solved first and the merge operation being recursively applied from the bottom to the top.
 
\subsection{Sequential tasks}\label{app:seq-tasks}
\begin{figure*}[!ht]
    \includegraphics[width=\textwidth]{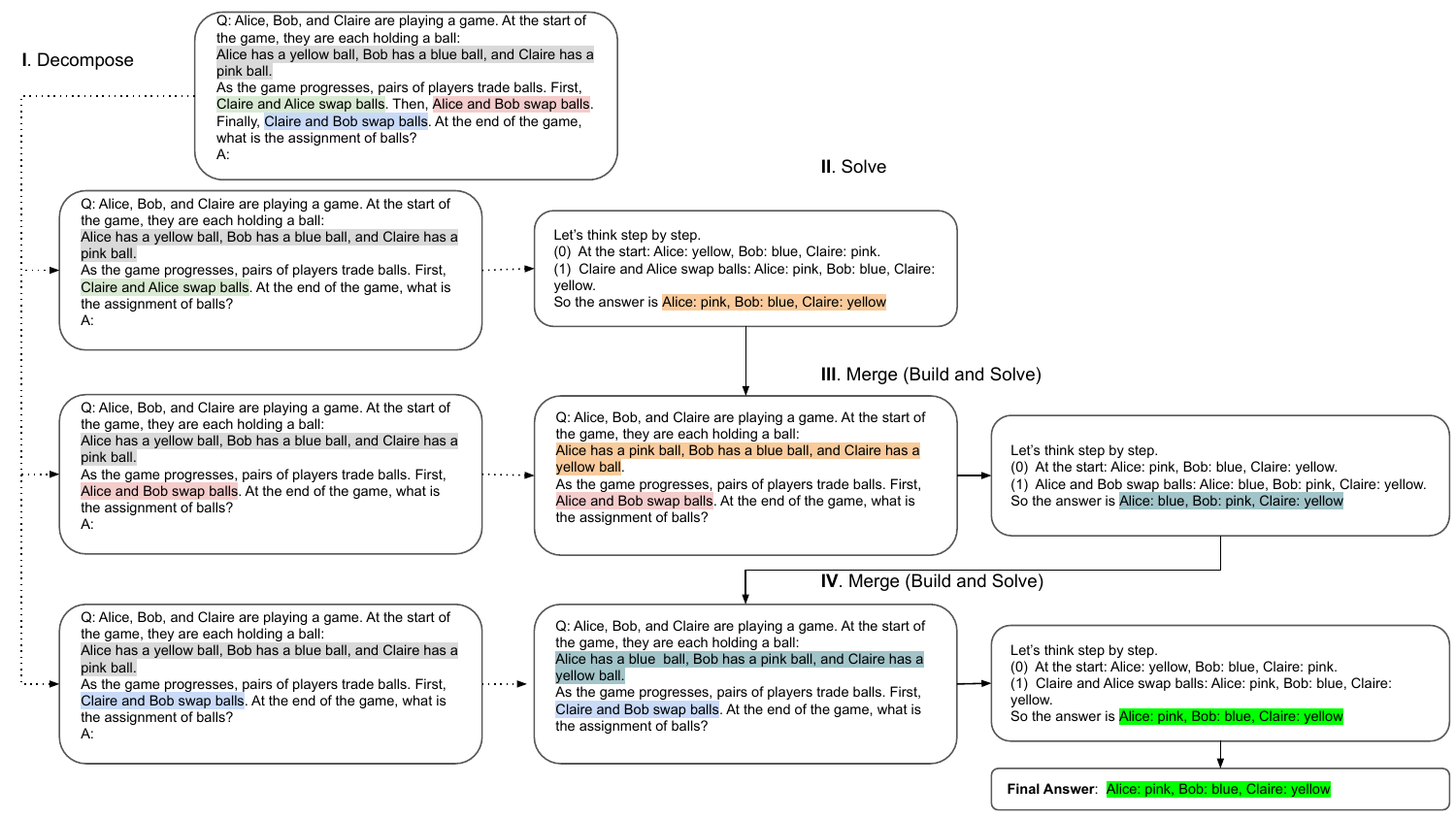}
    \caption{Overview of $ToP~(1, 3)$ on an instance of \textit{Tracking Shuffled Objects (three objects)}.}
    \label{fig:top_1_3}
\end{figure*}

Let us say that we have a system at state $s_0$, and we want to find its state after going through $m$ processing steps $(p_1, \ldots, p_m)$ in this order (i.e.~a sequential task). Applying $ToP~(1, k)$ is equivalent to grouping the above steps into $k$ groups $G_{1} = \left (p_1, \ldots, p_{\lceil \frac{m}{k}\rceil} \right ), \ldots, G_k = \left ( p_{m - \lfloor \frac{m}{k} \rfloor + 1}, \ldots, p_m \right)$. We build a path graph from top to bottom, where the root is the main instance, and the leaf is the instance defined by $s_0$ and $G_1$. Solving it yields a state $s_1$ to which we apply the steps $G_2$ and so on until we reach $G_k$. \textit{Tracking Shuffled Objects} is an example of such a task. At the start, $L$ people are assigned one object each. We are interested in recovering the assignment between people and objects after $L$ swaps (transpositions). Figure~\ref{fig:top_1_3} illustrates the application of $ToP~(1, 3)$ to an instance with 3 swaps. We first decompose the main instance into $3$ subinstances; here, each instance corresponds to one swap. After decomposition, only the first instance has the correct initial assignment (grey part). For the remaining instances, placeholders are used, which will later be replaced by the solutions to the problems they depend on.

\subsection{Comparison with Least-to-Most Prompting}\label{app:L2M}
\begin{figure*}[!ht]
    \includegraphics[width=\textwidth]{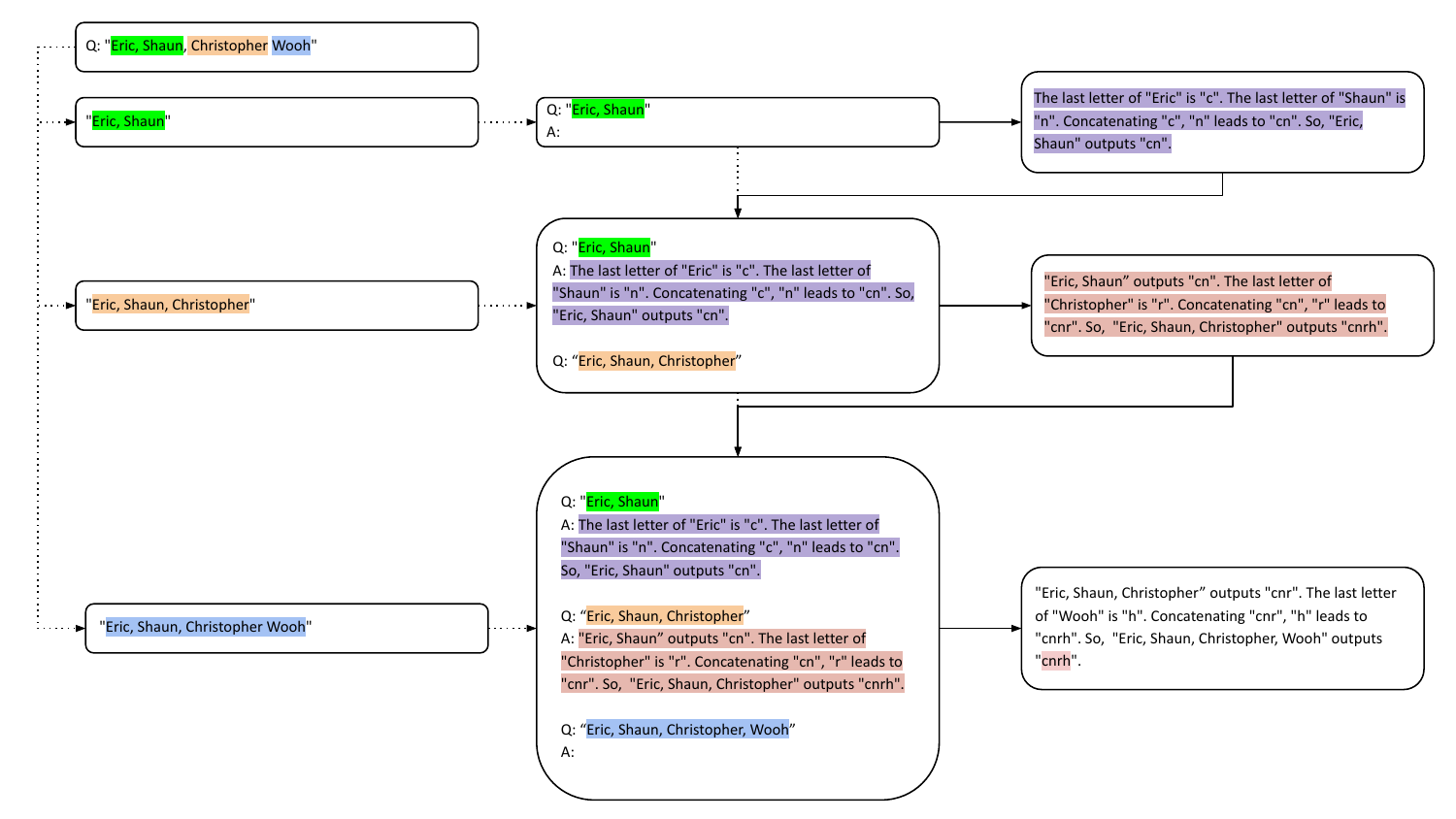}
    \caption{Overview of L2M prompting on Last Letter Concatenation with 4 words.}
    \label{fig:l2m_4}
\end{figure*}

Least-to-Most prompting also handles Last Letter Concatenation as a sequential task. In this regards, it is similar to $ToP~(1, L)$ on list with $L$ words. As illustrated in Figure~\ref{fig:l2m_4}, L2M uses all couples instance-solution preceding an instance to build the prompt to solve it whereas ToP only uses the couples directly connected to it in the tree hierarchy.

\section{Additional Experiments} \label{section:additional}

\subsection{Scaling behaviour} \label{appendix:scaling}
\begin{figure*}[t]
\centering
%\begin{figure}[t]
%\includegraphics[width=\columnwidth{figures/scale.pdf}
  \includegraphics[width=0.9\textwidth]{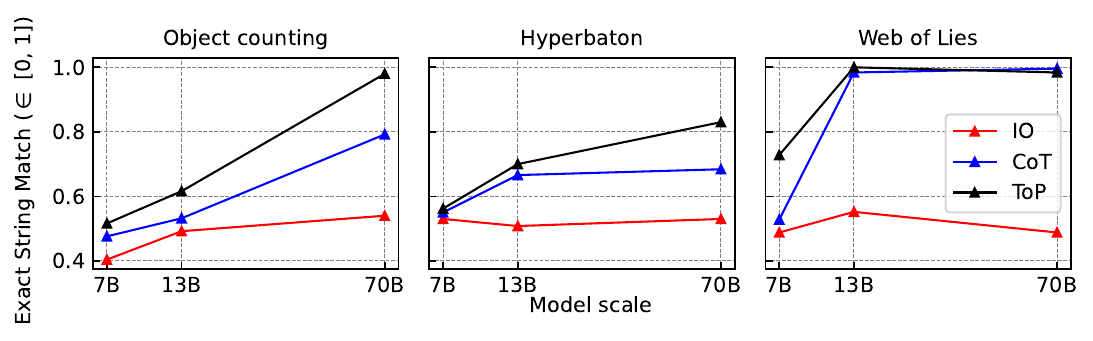}
  \caption{Scaling behavior of ToP compared to IO and CoT with the LLaMA 2 family on 3 BBH tasks.}
  \label{fig:scaling}
%\end{figure}
\end{figure*}

%\paragraph{Scaling behaviour}
In this section, we study how ToP behaves as we vary the model scale. In Figure~\ref{fig:scaling}, we plot the performance of both IO and CoT prompting and ToP as a function of model scale for LLaMA~2 models~\citep{touvron2023llama} and 3 BBH tasks.  We use ToP~(2, 1) for canonical tasks and ToP~(1, 2) for sequential tasks. For all tasks, scaling up model size improved the performance of ToP beyond CoT prompting. LLaMA~2~70B achieves a 98\% accuracy on \textit{Object Counting}, an absolute improvement of 18.8\% over CoT. ToP improves over random accuracy of IO and CoT on \textit{Web of Lies} with LLaMA 2 7B, with an accuracy of 72.8\%.

We report IO prompting, CoT prompting and ToP performance on 8 BBH tasks in Table~\ref{tab:bbh_llama_2}. ToP consistently yields an improvement of performance compared to IO and CoT prompting for most tasks and at all scales.

\begin{table*}[ht]
%\tiny
\small
\begin{center}
\resizebox{\textwidth}{!}{
%\resizebox{\columnwidth}{!}{
\begin{tabular}{lrrrrrrrrrrrr}
\toprule
\multirow{2}{*}{\bf BBH Tasks}  &\multicolumn{3}{c}{\bf LLaMA 2 7B} & \multicolumn{1}{c}{\bf } &\multicolumn{3}{c}{\bf LLaMA 2 13B} & \multicolumn{1}{c}{\bf } & \multicolumn{3}{c}{\bf LLaMA 2 70B} \\
\cmidrule{2-4} \cmidrule{6-8} \cmidrule{10-12}
{} & IO & CoT & ToP & & IO & CoT & ToP & & IO & CoT & ToP \\
\midrule
\textit{Boolean Expressions} & 0.680 & 0.628 & \bf 0.672 & & 0.728 & \bf 0.768 & 0.728 & & 0.812 & 0.868 & \bf 0.924\\
\textit{Hyperbaton} & 0.530 & 0.550 & \bf 0.562 & & 0.508 & 0.666 & \bf 0.700 & & 0.530 & 0.684 & \bf 0.830 \\
%\midrule
\textit{Multistep Arithmetic Two} & 0.008 & 0.004 & \bf 0.012 & & 0.012 & 0.024 & \bf 0.044 & & 0.016 & 0.196 & \bf 0.216 \\
\textit{Navigate} & \bf 0.272 & 0.164 & 0.088 & & \bf 0.340 & 0.308 & 0.156 & & 0.336 & \bf 0.400 & 0.284 \\
\textit{Object Counting} & 0.404 & 0.476 & \bf 0.516 & & 0.492 & 0.532 & \bf 0.616 & & 0.540 & 0.792 & \bf 0.98 \\
\midrule
\textit{Tracking Shuffled Objects} \\
\midrule
\textit{Three} & \bf 0.156 & \bf 0.156 & 0.136 & & 0.076 & \bf 0.184 & 0.132 & & 0.056 & \bf 0.584 & 0.568 \\
\textit{Five} & 0.000 & 0.000 & 0.000 & & 0.012 & 0.044 & \bf 0.048 & & 0.080 & 0.528 & \bf 0.664 \\
\textit{Seven} & 0.000 & 0.000 & 0.000 & & 0.000 & 0.000 & \bf 0.004 & & 0.000 & 0.288 & \bf 0.592 \\
\midrule
\textit{Web of Lies} & 0.488 & 0.528 & \bf 0.728 & & 0.552 & 0.984 & \bf 1.000 & & 0.488 & \bf 0.996 & 0.984 \\
\textit{Word Sorting} & \bf 0.418 & 0.146 & 0.244 & & \bf 0.538 & 0.261 & 0.320 & & \bf 0.788 & 0.445 & 0.717\\
\bottomrule
\end{tabular}
}
\end{center}
\caption{Few-shot prompting performance of the LLaMA 2 family on BIG-Bench Hard (BBH).}
\label{tab:bbh_llama_2}
\end{table*}

\section{Analysis} \label{appendix:analysis}
We aim to gain a comprehensive understanding of the performance improvements offered by the ToP framework. We theoretically derive an upper bound of expected ToP's performance, then we study the impact of the tree structure on the results obtained. For the experiments in this section, we use LLaMA~3~8B \citep{dubey2024llama3herdmodels} unless stated otherwise.

\subsection{Theoretical Analysis} \label{appendix:theory}
\begin{figure}[t]
  \includegraphics[width=\columnwidth]{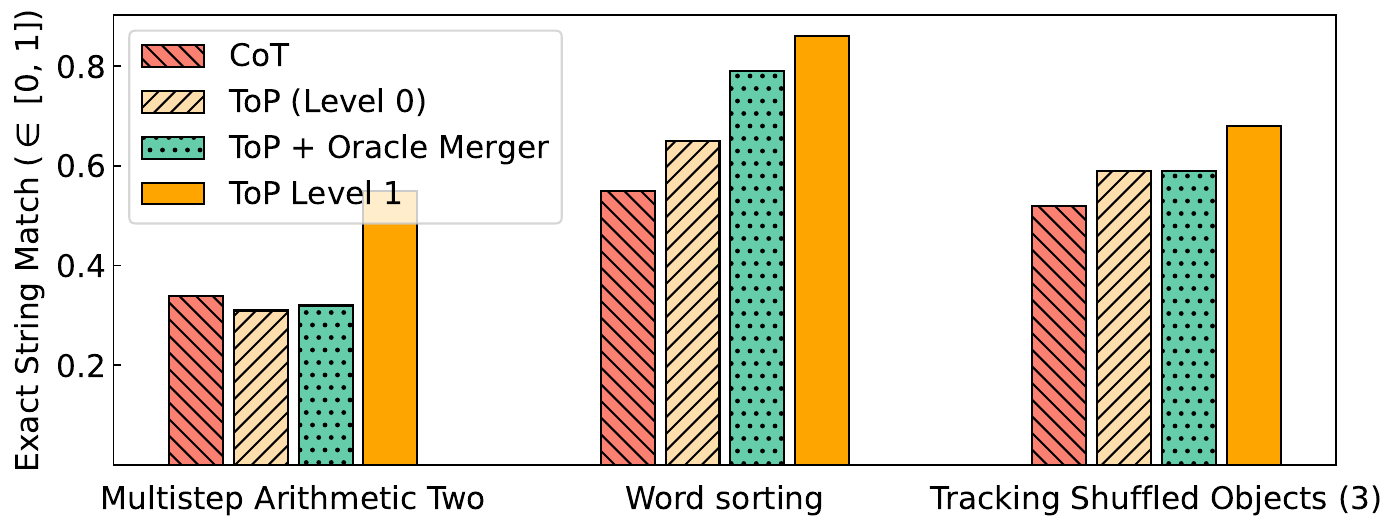}
  \caption{Comparison between CoT, ToP, ToP with an Oracle Merger and the leaves' accuracy on 3 BBH tasks.}
  \label{fig:comparison}
\end{figure}

Let us consider a task with $n$ problems. Each problem is further divided into $k$ subproblems, resulting in a total of $nk$ subproblems. If we evaluate an LLM on these $nk$ subproblems and obtain $m$ incorrect answers, we can infer the number of incorrect answers likely to occur when evaluating the original $n$ problems. Assuming that an incorrect answer to a subproblem implies an incorrect answer to its corresponding main problem, we can analyze the outcomes in two scenarios. In the worst case, each of the $m$ incorrect subproblems is associated with a distinct main problem and thus there would be $m$ main problems with incorrect answers. The best case is when the $m$ incorrect subproblems are distributed such that each affected main problem has $k$ or $m\%k$ incorrect subproblems. Consequently, the number of main problems with incorrect answers would be at most $\lceil \frac{m}{k} \rceil$. From this analysis, we can deduce that the accuracy at any level $l$ of the problem hierarchy is constrained by the accuracy at level $l$ - 1. Therefore, the accuracy for the overall task (the root of the hierarchy) is bounded by the accuracy observed at the most granular level (the leaves of the hierarchy). We validate this analysis by comparing the accuracy at level 1 to the accuracy at level 0 (main problem) for some of the aforementioned BBH tasks. The results are summarized in Figure~\ref{fig:comparison}. The Oracle Merger represents the accuracy that would be achieved if the merger process were flawless.

As expected, the accuracy at the leaves acts as an upper bound for ToP. Moreover, the Oracle Merger yields better performance than vanilla ToP. This suggests that there is a loss in accuracy when going from level $k$ to level $k - 1$, which can prevent ToP from achieving an even higher performance. Interestingly, what happens with \textit{Multistep Arithmetic Two} comes close to the worst case scenario that we depicted earlier. Despite the leaves' accuracy being 55\%, ToP + Oracle Merger fails to outperform CoT's 34\% accuracy, showing that the distribution of the correct leaves' instances inherently undermines ToP performance in this scenario.

\subsection{Impact of the tree structure.}\label{appendix:structure}
\paragraph{GoT Tasks.} 
We analyze the impact of the tree structure on ToP's results. As shown previously, there may be a loss in accuracy during the merge operation. A deeper tree means more of these losses, but it also means easier subproblems. For the three GoT tasks, we analyze the impact of the tree's depth when the breadth is set to two with LLaMA~3~70B~Instruct~\citep{dubey2024llama3herdmodels}.

\begin{figure}[t]
\centering
  \includegraphics[width=\columnwidth]{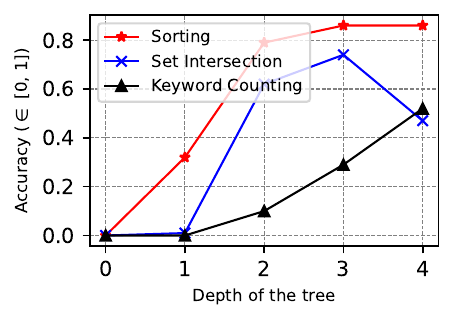}
  \caption{Impact of the tree structure (depth) on the accuracy on the 3 \textbf{GoT Tasks} with LLaMA~3~70B~Instruct. Depth = 0 represents CoT prompting.}
  \label{fig:structure_got}
\end{figure}

As suggested by Figure~\ref{fig:structure_got}, deeper trees led to a higher accuracy for all three tasks. This is because we observed very few errors during the merge operation performed by the LLM. Going deeper, even with a near perfect merger can negatively affect performance as observed with the Set Intersection task, which has an accuracy of 47\% with $d = 4$ but 74\% with $d = 3$ and 62\% with $d = 2$. The small errors performed at the leaves being propagated during the repetitive merge operations impact the overall accuracy of ToP. In terms of breadth, applying ToP~(4, 1) to \textit{Set Intersection} yields the same accuracy of 62\% as ToP~(2, 2). We observed ToP~(4, 2) to have a 49\% accuracy, comparable to ToP~(2, 4)'s 47\%.

\begin{figure}[t]
\centering
  \includegraphics[width=\columnwidth]{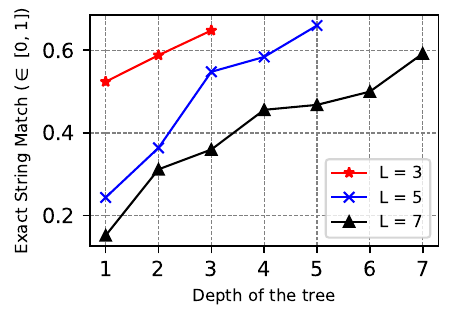}
  \caption{Impact of the tree structure (depth) on the accuracy of ToP on \textit{Tracking Shuffled Objects $\left ( L \in \{3, 5, 7\} \right )$}. Depth = 0 and depth = 1 represent CoT prompting.}
  \label{fig:structure}
\end{figure}

\paragraph{BBH Tasks.}
\textit{Tracking Shuffled Objects} involves recovering the final assignement of $L$ objects given to $L$ people $\left ( L \in \{3, 5, 7\} \right )$ after a series of $L$ transpositions (pairwise swaps). Applying ToP~(1, $d$) to these tasks implies using $d$ even subseries of swaps in a manner akin to \textit{Navigate} (see Figure~\ref{fig:fig1}). We study the impact of various depths and we report the results in Figure~\ref{fig:structure}.

Across all settings, the task accuracy gradually increases with deeper trees and reaches its maximum when all the subproblems involve only one swap (depth = $L - 1$). The trade-off between the number of merge operations and the accuracy of simple instances is not at play here.

On \textit{Multistep Arithmetic Two}, ToP~(2, 1) and ToP~(2, 2) respectively achieve 30.8\% and 57.2\% accuracy where CoT is at 34\%. Similarly, ToP~(1, 2) and ToP~(1, 3) respectively achieve 60\% and 66.4\% where CoT is at 60.4\% on \textit{Navigate}. This suggests that the tree structure can greatly affect the quality of ToP.

\subsection{Robustness to the solve prompt.} \label{appendix:robustness}

Throughout our experiments, we used CoT prompting to solve the most granular subproblems (the tree's leaves). In this section, we examine the impact of using IO prompting to solve the leaves. We conduct experiments on $\textit{Word Sorting}$, which did not benefit from CoT prompting as shown in Table~\ref{tab:bbh_canonical}. Additionally, we include \textit{Tracking Shuffled Objects (3, 5)}, \textit{Boolean Expressions}, \textit{Multistep Arithmetic Two}, and \textit{Object Counting}, where IO prompting produced much poorer results compared to CoT. The results are summarized in Table~\ref{tab:robustness}.

\begin{table}[ht]
\centering\small
\begin{center}
\resizebox{\columnwidth}{!}{
\begin{tabular}{lrrrr}
\toprule
\multirow{2}{*}{\bf BBH tasks}  & \multicolumn{2}{c}{\bf LLaMA~3~8B} \\
\cmidrule{2-3}
{} & IO & IO + ToP \\
\midrule
\textit{Boolean Expressions} & 0.824 & \bf 0.876 \\
\textit{Multistep Arithmetic Two} & 0.008 & \bf 0.036 \\
\textit{Object Counting} & 0.492 & \bf 0.552 \\
\midrule
\textit{Tracking Shuffled Objects} \\
\midrule
\textit{Three} & 0.132 & \bf 0.196 \\
\textit{Five} & 0.004 & \bf 0.008 \\
\midrule
\textit{Web of Lies} & \bf 0.528 & 0.484 \\
\textit{Word Sorting} & 0.647 & \bf 0.679 \\
\bottomrule
\end{tabular}
}
\end{center}
\caption{Comparison of IO prompting and IO + ToP on BBH tasks.}
\label{tab:robustness}
\end{table}

We observe that IO + ToP consistently outperforms standalone IO prompting on all the BBH tasks except \textit{Web of Lies}. This is because the LLM's performance is random (0.528) on that specific task and dividing it into simpler instances does not resolve the issue (leaves' EM = 0.496). Moreover, applying IO + ToP on \textit{Multistep Arithmetic Two} with a depth of 2 instead of 1 gives a 22\% accuracy. These results demonstrate that ToP is robust to low-quality \verb|solve_prompt| but achieves the best results with a good \verb|solve_prompt|.

\subsection{Analysis of failure cases} \label{appendix:failure_cases}
In Table~\ref{tab:bbh_canonical}, we showed that ToP induces considerable performance gains on various complex BBH tasks with \texttt{gpt-3.5-turbo-instruct}. However, ToP specifically failed to improve performance on \textit{Multistep Arithmetic Two} and \textit{Object Counting}. An analysis of the model's outputs suggests 2 main causes:

\paragraph{1. The founding hypothesis of ToP is not always verified.} The founding hypothesis of ToP is the ability of an LLM to perform better on simpler instances of a task compared to more complex ones. When this is not verified, ToP's performance tends to fall behind CoT prompting. We strikingly observe such a behaviour on \textit{Object Counting}. For multiple instances, the LLM was able to count the relevant items in a given list but fails to do so in a subset of the same list, as illustrated in Table~\ref{tab:object_error}. This also frequently occurs with \textit{Multistep Arithmetic Two} instances, where the 73.6\% ToP's accuracy cannot be improved even with an Oracle merger.

\begin{table}[ht]
%\tiny
\small
\begin{center}
%\begin{tabular}{l}
\begin{tabular}{p{7cm}}
\toprule
\textbf{\textsc{Main instance}} \\
\midrule
\textbf{Q}: I have three oranges, a pig, a frog, a cow, three bananas, a nectarine, and a snail. How many animals do I have?\\
\midrule
Let's think step by step.\\
We first identify the animals on the list and include their quantity in parentheses:\\
- pig (1)\\
- frog (1)\\
- cow (1)\\
- snail (1)\\
Now, let's add the numbers in parentheses: 1 + 1 + 1 + 1 = 4. So the answer is 4. \\
\midrule
\textbf{\textsc{Subinstance}} \\
\midrule
\textbf{Q}: I have three bananas, a nectarine, and a snail. How many animals do I have? \\
\midrule
Let's think step by step.\\
We first identify the animals on the list and include their quantity in parentheses:\\
\textcolor{red}{- bananas (3)} \\
\textcolor{red}{- nectarine (1)} \\ 
- snail (1)\\
Now, let's add the numbers in parentheses: 3 + 1 + 1 = 5. So the answer is 5. \\
\bottomrule
\end{tabular}
\end{center}
\caption{\texttt{gpt-3.5-turbo-instruct} is able to accurately count the number of animals in a given list but fails to count it in a sublist.}
\label{tab:object_error}
\end{table}

\paragraph{2. The LLM fails to merge subproblems' solutions into their parent's solution.} In Appendix~\ref{appendix:theory}, we showed that the use of an oracle merger leads to even greater improvements in the ToP framework. Additionally, a bad merger can bring ToP below CoT performance. Table~\ref{tab:object_merge_error} shows an example on the \textit{Object Counting} task, where the LLM accurately solve both subproblems but fail to combine their answers. We believe that such issues can be significantly alleviated with a better \verb|merge_prompt|.

\begin{table}[!ht]
%\tiny
\small
\begin{center}
%\begin{tabular}{l}
\begin{tabular}{p{7cm}}
\toprule
\textbf{\textsc{Subinstance 1}} \\
\midrule
\textbf{Q}: I have a yam, a cauliflower, a bed, two cabbages, a garlic, and an oven. How many vegetables do I have?\\
\midrule
Let's think step by step.\\
We first identify the vegetables on the list and include their quantity in parentheses:\\
- yam (1)\\
- cauliflower (1)\\
- cabbage (2)\\
- garlic (1)\\
Now, let's add the numbers in parentheses: 1 + 1 + 2 + 1 = 5. So the answer is 5.\\
\midrule
\textbf{\textsc{Subinstance 2}} \\
\midrule
\textbf{Q}: I have a carrot, a head of broccoli, a potato, a stalk of celery, a lettuce head, and a toaster. How many vegetables do I have? \\
\midrule
Let's think step by step.\\
We first identify the vegetables on the list and include their quantity in parentheses:\\
- carrot (1)\\
- broccoli (1)\\
- potato (1)\\
- celery (1)\\
- lettuce (1)\\
Now, let's add the numbers in parentheses: 1 + 1 + 1 + 1 + 1 = 5. So the answer is 5.\\
\midrule
\textbf{\textsc{Main instance}} \\
\midrule
\textbf{Q}: I have a yam, a cauliflower, a bed, two cabbages, a garlic, an oven, a carrot, a head of broccoli, a potato, a stalk of celery, a lettuce head, and a toaster. How many vegetables do I have? \\
\midrule
Let's think step by step. \\
The list of words can be written as the concatenation of A and B where \\
- A = a yam, a cauliflower, a bed, two cabbages, and a garlic.\\
- B = \textcolor{red}{an oven}, a carrot, a head of broccoli, a potato, a stalk of celery, and a lettuce head.\\
According to the premises, A contains 5 vegetables.\\
According to the premises, B contains \textcolor{red}{6} vegetables.\\
Then, the total number of vegetables is A + B = 5 + \textcolor{red}{6} = \textcolor{red}{11}. So the answer is \textcolor{red}{11}. \\
\bottomrule
\end{tabular}
\end{center}
\caption{\texttt{gpt-3.5-turbo-instruct} is unable to use the information in its context to solve the given problem.}
\label{tab:object_merge_error}
\end{table}

\section{Implementation details} \label{appendix:details}

\subsection{Language Models and Datasets} \label{appendix:LLMandDS}

In Table~\ref{tab:url}, we list the links to the relevant resources used to build this work.
\begin{table*}[!ht]
%\begin{table}[!ht]
    \centering
    \resizebox{\linewidth}{!}{
    \begin{tabular}{l l}
    \toprule
    %Name & Public Link or Endpoint\\
    %\midrule
    \multicolumn{2}{c}{\textit{Datasets for Comparisons}} \\
    \midrule
    BIG-Bench Hard & \url{https://huggingface.co/datasets/lukaemon/bbh} \\
    \midrule
    \multicolumn{2}{c}{\textit{Models for Evaluations}} \\
    \midrule
    %Babbage-002 & babbage-002\\
    %Davinci-002 & davinci-002 \\
    %GPT-3.5-Turbo & gpt-3.5-turbo-0125	\\
    GPT-3.5-Turbo & gpt-3.5-turbo (gpt-3.5-turbo-0125)	\\
    GPT-3.5-Turbo-Instruct & gpt-3.5-turbo-instruct	\\
    LLaMA 3 8B & \url{https://huggingface.co/meta-llama/Meta-Llama-3-8B} \\
    LLaMA 3 70B & \url{https://huggingface.co/casperhansen/llama-3-70b-instruct-awq} \\
    LLaMA 2 7B & \url{https://huggingface.co/meta-llama/Llama-2-7b-hf} \\
    LLaMA 2 13B & \url{https://huggingface.co/meta-llama/Llama-2-13b-hf} \\
    LLaMA 2 70B & \url{https://huggingface.co/TheBloke/Llama-2-70B-AWQ} \\
    \bottomrule
    \end{tabular}}
    \caption{Links to datasets, benchmarks and language models.}
    \label{tab:url}
%\end{table}
\end{table*}

\subsection{Sampling scheme and evaluation metrics}
For prompting strategies that require sampling, we use nucleus sampling with $T = 0.7$ and $\textrm{top-p} = 0.95$. Otherwise, we use greedy decoding ($T = 0.0$). In all experiments, we generate at most 2000 new tokens and use Exact String Match (EM) as the evaluation metric.

\subsection{Decomposition}
All the decomposers are algorithmic. For sorting and set intersection, a list \texttt{L} is divided into \texttt{L[0:len(L)//2]} and \texttt{L[len(L)//2:]}. For keyword sorting, we decompose the input text based on its sentences with the help of a sentence splitter module.\footnote{\url{https://github.com/mediacloud/sentence-splitter}}

\subsection{How to choose the breadth and the depth}
ToP depends on 2 parameters, the breadth and the depth of the tree structure. A quick analysis of the problem can lead to an informed guess about what a good breadth should be. This is typically the case of sorting problems when a breadth of 2 helps to mimic the merge sort algorithm. We mostly experimented with a breadth of 2 for canonical tasks and saw that it yielded very good results. When it comes to sequential problems, the breadth is 1 and the depth plays the role of the number of block of steps before reaching the final state (depth-wise decomposition). Using 2 blocks also gave good results, but deeper trees tend to always give better results for such problems.

\section{Prompts} \label{app:prompts}

\subsection{GoT Tasks}
We provide the links to all the prompts used to solve the GoT tasks in Table~\ref{tab:got_prompts}.

\begin{table*}[!ht]
%\begin{table}[!ht]
    \centering
    \resizebox{\linewidth}{!}{
    \begin{tabular}{l l}
    \toprule
    \multicolumn{2}{c}{\textit{CoT}} \\
    \midrule
    Sorting & \url{https://github.com/ArmelRandy/tree-of-problems/blob/master/top/got/prompts/cot/sorting.txt}\\
    Set Intersection & \url{https://github.com/ArmelRandy/tree-of-problems/blob/master/top/got/prompts/cot/set_intersection.txt}\\
    Keyword Counting & \url{https://github.com/ArmelRandy/tree-of-problems/blob/master/top/got/prompts/cot/keyword_counting.txt}\\
    \midrule
    \multicolumn{2}{c}{\textit{Merge}} \\
    \midrule
    Sorting & \url{https://github.com/ArmelRandy/tree-of-problems/blob/master/top/got/prompts/merge/sorting.txt}\\
    Set Intersection (2) & \url{https://github.com/ArmelRandy/tree-of-problems/blob/master/top/got/prompts/merge/set_intersection.txt}\\
    Set Intersection (4) & \url{https://github.com/ArmelRandy/tree-of-problems/blob/master/top/got/prompts/merge/set_intersection_4.txt}\\
    Keyword Counting & \url{https://github.com/ArmelRandy/tree-of-problems/blob/master/top/got/prompts/merge/keyword_counting.txt}\\
    \bottomrule
    \end{tabular}}
    \caption{Links to solve and merge prompts of the GoT Tasks.}
    \label{tab:got_prompts}
%\end{table}
\end{table*}

\subsection{BBH tasks}\label{section:modification}

We describe the modification applied to 3 BBH tasks: \textit{Hyperbaton}, \textit{Navigate} and \textit{Tracking Shuffled Objects}. Instead of choosing which of two sentences has the right adjective ordering, we turned each instance of \textit{Hyperbaton} into 2 separate instances of Yes/No questions relative to adjectives ordering. Table~\ref{tab:hyperbaton} shows an example instance before and after the modification (we only reported one separate instance for visibility). We modify \textit{Navigate} to require to find the final position after a series of instructions instead of asking if they lead to the starting point. Table~\ref{tab:navigate} shows an example instance before and after modification. In \textit{Tracking Shuffled Objects}, we ask the LLM to recover the final object assignments after the pairwise swaps as exemplified in Table~\ref{tab:tso}.

\begin{table*}[ht]
%\tiny
\small
\begin{center}
%\begin{tabular}{l}
\begin{tabular}{p{15.5cm}}
\toprule
\textbf{\textsc{Before}} \\
\midrule
\textbf{Q}: Which sentence has the correct adjective order: \\
Options: \\
(A) rubber terrible ship \\
(B) terrible rubber ship \\
\midrule
IO Answer \\
\midrule
\textbf{(B)} \\
\midrule
CoT Answer \\
\midrule
\textbf{A}: Let's think step by step. \\
When there is more than one adjective before a noun, the adjectives need to respect the following order before a noun: "[1.
opinion] [2. size] [3. age] [4. shape] [5. color] [6. origin] [7. material] [8. purpose] noun". \\
Option (A): "rubber terrible ship". (1) rubber" falls into the material category. (2) "terrible" falls into the opinion category. Option (A)
has the following adjective order: [7. material] [1. opinion] (or, in numeric terms, 7 1). Because 7 < 1 is not correct, (A) does not have
the correct ordering. \\
Option (B): "terrible rubber ship". Option (B) has the following adjective order: [1. opinion] [7. material] (or, in numeric terms, 1 7).
Because 1 < 7 is correct, (B) has the correct ordering. So the answer is \textbf{(B)}. \\
\\
\midrule
\textbf{\textsc{After}} \\
\midrule
\textbf{Q}: \textcolor{blue}{Answer with Yes or No. Does the following sentence have the correct adjective order?} \\
rubber terrible ship \\
\midrule
IO Answer \\
\midrule
\textbf{No} \\
\midrule
CoT Answer \\
\midrule
\textbf{A}: Let's think step by step. \\
When there is more than one adjective before a noun, the adjectives need to respect the following order before a noun: "[1.
opinion] [2. size] [3. age] [4. shape] [5. color] [6. origin] [7. material] [8. purpose] noun". \\
The sentence is "rubber terrible ship". (1) rubber" falls into the material category. (2) "terrible" falls into the opinion category. \\
The sentence has the following adjective order: [7. material] [1. opinion] (or, in numeric terms, 7 1). Because 7 < 1 is not correct, the sentence does not have
the correct ordering. So the answer is \textbf{No}. \\
\bottomrule
\end{tabular}
\end{center}
\caption{Modification applied to \textit{Hyperbaton}.}
\label{tab:hyperbaton}
\end{table*}

\begin{table*}[ht]
%\tiny
\small
\begin{center}
%\begin{tabular}{l}
\begin{tabular}{p{15.5cm}}
\toprule
\textbf{\textsc{Before}} \\
\midrule
\textbf{Q}: If you follow these instructions, do you return to the starting point? Turn left. Turn around. Turn left. Take 7 steps. Take 2 steps. Take 4 steps. Take 8 steps. \\
\midrule
IO Answer \\
\midrule
\textbf{No} \\
\midrule
CoT Answer \\
\midrule
\textbf{A}: Let's think step by step.
We start at the origin (0, 0), facing the positive y-axis. \\
(1) Turn left: (0, 0), facing the negative x-axis. \\
(2) Turn around: (0, 0), facing the positive x-axis. \\
(3) Turn left: (0, 0), facing the positive y-axis. \\
(4) Take 7 steps: (0, 7), facing the positive y-axis. \\
(5) Take 2 steps: (0, 9), facing the positive y-axis. \\
(6) Take 4 steps: (0, 13), facing the positive y-axis. \\
(7) Take 8 steps: (0, 21), facing the positive y-axis. \\
Since (0, 21) is not (0, 0), we are not where we started. So the answer is \textbf{No}. \\
\\
\midrule
\textbf{\textsc{After}} \\
\midrule
\textbf{Q}: If you follow these instructions, \textcolor{blue}{what are the coordinates of the end point if you start at the point (0, 0), facing the positive y-axis?} Turn left. Turn around. Turn left. Take 7 steps. Take 2 steps. Take 4 steps. Take 8 steps. \\
\midrule
IO Answer \\
\midrule
\textbf{(0, 21)} \\
\midrule
CoT Answer \\
\midrule
\textbf{A}: Let's think step by step.
We start at the origin (0, 0), facing the positive y-axis. \\
(1) Turn left: (0, 0), facing the negative x-axis. \\
(2) Turn around: (0, 0), facing the positive x-axis. \\
(3) Turn left: (0, 0), facing the positive y-axis. \\
(4) Take 7 steps: (0, 7), facing the positive y-axis. \\
(5) Take 2 steps: (0, 9), facing the positive y-axis. \\
(6) Take 4 steps: (0, 13), facing the positive y-axis. \\
(7) Take 8 steps: (0, 21), facing the positive y-axis. \\
So the answer is \textbf{(0, 21)}. \\
\bottomrule
\end{tabular}
\end{center}
\caption{Modification applied to \textit{Navigate}.}
\label{tab:navigate}
\end{table*}

\begin{table*}[ht]
%\tiny
\small
\begin{center}
%\begin{tabular}{l}
\begin{tabular}{p{15.5cm}}
\toprule
\textbf{\textsc{Before}} \\
\midrule
\textbf{Q}: Alice, Bob, and Claire are friends and avid readers who occasionally trade books. At the start of the semester, they each buy one new book: Alice gets Ulysses, Bob gets Frankenstein, and Claire gets Lolita. As the semester proceeds, they start trading around the new books. First, Claire and Bob swap books. Then, Bob and Alice swap books. Finally, Claire and Bob swap books. \\
At the end of the semester, Bob has  \\
Options: \\
(A) Ulysses \\
(B) Frankenstein \\ 
(C) Lolita \\
\midrule
IO Answer \\
\midrule
\textbf{(B)} \\
\midrule
CoT Answer \\
\midrule
A: Let's think step by step. \\
(0) At the start: Alice: Ulysses, Bob: Frankenstein, Claire: Lolita. \\
(1) Claire and Bob swap books: Alice: Ulysses, Bob: Lolita, Claire: Frankenstein. \\
(2) Bob and Alice swap books: Alice: Lolita, Bob: Ulysses, Claire: Frankenstein. \\
(3) Claire and Bob swap books: Alice: Lolita, Bob: Frankenstein, Claire: Ulysses.  \\
At the end of the semester, Bob has Frankenstein. So the answer is \textbf{(B)}.
\\
\midrule
\textbf{\textsc{After}} \\
\midrule
\textbf{Q}: Alice, Bob, and Claire are friends and avid readers who occasionally trade books. At the start of the semester, they each buy one new book: Alice gets Ulysses, Bob gets Frankenstein, and Claire gets Lolita. As the semester proceeds, they start trading around the new books. First, Claire and Bob swap books. Then, Bob and Alice swap books. Finally, Claire and Bob swap books. \\
At the end of the semester, \textcolor{blue}{what is the assignment of books?} \\
\midrule
IO Answer \\
\midrule
\textbf{Alice: Lolita, Bob: Frankenstein, Claire: Ulysses} \\
\midrule
CoT Answer \\
\midrule
\textbf{A}: Let's think step by step. \\
(0) At the start: Alice: Ulysses, Bob: Frankenstein, Claire: Lolita. \\
(1) Claire and Bob swap books: Alice: Ulysses, Bob: Lolita, Claire: Frankenstein. \\
(2) Bob and Alice swap books: Alice: Lolita, Bob: Ulysses, Claire: Frankenstein. \\
(3) Claire and Bob swap books: Alice: Lolita, Bob: Frankenstein, Claire: Ulysses.  \\
So the answer is \textbf{Alice: Lolita, Bob: Frankenstein, Claire: Ulysses}. \\
\bottomrule
\end{tabular}
\end{center}
\caption{Modification applied to \textit{Tracking Shuffled Objects (Three objects)}.}
\label{tab:tso}
\end{table*}

We provide the links to all the prompts used to solve the BBH tasks in Table~\ref{tab:bbh_prompts}.

\begin{table*}[!ht]
%\begin{table}[!ht]
    \centering
    \resizebox{\linewidth}{!}{
    \begin{tabular}{l l}
    \toprule
    %Name & Public Link or Endpoint\\
    %\midrule
    \multicolumn{2}{c}{\textit{CoT}} \\
    \midrule
    Boolean Expressions & \url{https://github.com/ArmelRandy/tree-of-problems/blob/master/top/bbh/prompts/cot/boolean_expressions.txt}\\
    Hyperbaton & \url{https://github.com/ArmelRandy/tree-of-problems/blob/master/top/bbh/prompts/cot/hyperbaton.txt}\\
    Multistep Arithmetic Two & \url{https://github.com/ArmelRandy/tree-of-problems/blob/master/top/bbh/prompts/cot/multistep_arithmetic_two.txt}\\
    Navigate & \url{https://github.com/ArmelRandy/tree-of-problems/blob/master/top/bbh/prompts/cot/navigate.txt}\\
    Object Counting & \url{https://github.com/ArmelRandy/tree-of-problems/blob/master/top/bbh/prompts/cot/object_counting.txt}\\
    Tracking Shuffled Objects & \url{https://github.com/ArmelRandy/tree-of-problems/blob/master/top/bbh/prompts/cot/tracking_shuffled_objects.txt}\\
    Web of Lies & \url{https://github.com/ArmelRandy/tree-of-problems/blob/master/top/bbh/prompts/cot/web_of_lies.txt}\\
    Word Sorting & \url{https://github.com/ArmelRandy/tree-of-problems/blob/master/top/bbh/prompts/cot/word_sorting.txt}\\
    \midrule
    \multicolumn{2}{c}{\textit{IO}} \\
    \midrule
    Boolean Expressions & \url{https://github.com/ArmelRandy/tree-of-problems/blob/master/top/bbh/prompts/standard/boolean_expressions.txt}\\
    Hyperbaton & \url{https://github.com/ArmelRandy/tree-of-problems/blob/master/top/bbh/prompts/standard/hyperbaton.txt}\\
    Multistep Arithmetic Two & \url{https://github.com/ArmelRandy/tree-of-problems/blob/master/top/bbh/prompts/standard/multistep_arithmetic_two.txt}\\
    Navigate & \url{https://github.com/ArmelRandy/tree-of-problems/blob/master/top/bbh/prompts/standard/navigate.txt}\\
    Object Counting & \url{https://github.com/ArmelRandy/tree-of-problems/blob/master/top/bbh/prompts/standard/object_counting.txt}\\
    Tracking Shuffled Objects & \url{https://github.com/ArmelRandy/tree-of-problems/blob/master/top/bbh/prompts/standard/tracking_shuffled_objects.txt}\\
    Web of Lies & \url{https://github.com/ArmelRandy/tree-of-problems/blob/master/top/bbh/prompts/standard/web_of_lies.txt}\\
    Word Sorting & \url{https://github.com/ArmelRandy/tree-of-problems/blob/master/top/bbh/prompts/standard/word_sorting.txt}\\
    \midrule
    \multicolumn{2}{c}{\textit{Merge}} \\
    \midrule
    Boolean Expressions & \url{https://github.com/ArmelRandy/tree-of-problems/blob/master/top/bbh/prompts/merge/boolean_expressions.txt}\\
    Hyperbaton & \url{https://github.com/ArmelRandy/tree-of-problems/blob/master/top/bbh/prompts/merge/hyperbaton.txt}\\
    Multistep Arithmetic Two & \url{https://github.com/ArmelRandy/tree-of-problems/blob/master/top/bbh/prompts/merge/multistep_arithmetic_two.txt}\\
    Navigate & \url{https://github.com/ArmelRandy/tree-of-problems/blob/master/top/bbh/prompts/merge/navigate.txt}\\
    Object Counting & \url{https://github.com/ArmelRandy/tree-of-problems/blob/master/top/bbh/prompts/merge/object_counting.txt}\\
    Tracking Shuffled Objects & \url{https://github.com/ArmelRandy/tree-of-problems/blob/master/top/bbh/prompts/merge/tracking_shuffled_objects.txt}\\
    Web of Lies & \url{https://github.com/ArmelRandy/tree-of-problems/blob/master/top/bbh/prompts/merge/web_of_lies.txt}\\
    Word Sorting & \url{https://github.com/ArmelRandy/tree-of-problems/blob/master/top/bbh/prompts/merge/word_sorting.txt}\\
    \bottomrule
    \end{tabular}}
    \caption{Links to solve and merge prompts for the BBH tasks.}
    \label{tab:bbh_prompts}
%\end{table}
\end{table*}

%\subsection{Other Tasks}
\subsection{Symbolic Reasoning}

%We provide the links to all the prompts used to solve the Coin Flip, Last Letter Concatenation and Algebraïc sum in Table~\ref{tab:other_prompts}.
We provide the links to all the prompts used to solve Coin Flip and Last Letter Concatenation in Table~\ref{tab:other_prompts}.

\begin{table*}[!ht]
%\begin{table}[!ht]
    \centering
    \resizebox{\linewidth}{!}{
    \begin{tabular}{l l}
    \toprule
    \multicolumn{2}{c}{\textit{CoT}} \\
    \midrule
    %Algebraic & \url{https://github.com/ArmelRandy/tree-of-problems/blob/master/top/algebraic/prompts/cot/cot8.txt}\\
    Coin & \url{https://github.com/ArmelRandy/tree-of-problems/blob/master/top/coin/prompts/cot/cot8.txt}\\
    Concatenation & \url{https://github.com/ArmelRandy/tree-of-problems/blob/master/top/concatenation/prompts/cot/cot8.txt}\\
    \midrule
    \multicolumn{2}{c}{\textit{IO}} \\
    \midrule
    %Algebraic & \url{https://github.com/ArmelRandy/tree-of-problems/blob/master/top/algebraic/prompts/standard/standard8.txt}\\
    Coin & \url{https://github.com/ArmelRandy/tree-of-problems/blob/master/top/coin/prompts/standard/standard8.txt}\\
    Concatenation & \url{https://github.com/ArmelRandy/tree-of-problems/blob/master/top/concatenation/prompts/standard/standard8.txt}\\
    \midrule
    \multicolumn{2}{c}{\textit{Merge}} \\
    \midrule
    %Algebraic & \url{https://github.com/ArmelRandy/tree-of-problems/blob/master/top/algebraic/prompts/merge/merge.txt}\\
    Coin & \url{https://github.com/ArmelRandy/tree-of-problems/blob/master/top/coin/prompts/merge/merge.txt}\\
    Concatenation & \url{https://github.com/ArmelRandy/tree-of-problems/blob/master/top/concatenation/prompts/merge/merge.txt}\\
    \midrule
    \multicolumn{2}{c}{\textit{L2M}} \\
    \midrule
    %Algebraic & \url{https://github.com/ArmelRandy/tree-of-problems/blob/master/top/algebraic/prompts/merge/l2m.txt}\\
    Coin & \url{https://github.com/ArmelRandy/tree-of-problems/blob/master/top/coin/prompts/merge/l2m.txt}\\
    Concatenation & \url{https://github.com/ArmelRandy/tree-of-problems/blob/master/top/concatenation/prompts/merge/l2m.txt}\\
    \bottomrule
    \end{tabular}}
    %\caption{Links to solve and merge prompts for Coin Flip, Last Letter Concatenation and Algebraïc sum.}
    \caption{Links to solve and merge prompts for Coin Flip and Last Letter Concatenation.}
    \label{tab:other_prompts}
%\end{table}
\end{table*}

%/tree-of-problems/blob/master/top/ -> /tree-of-problems/blob/master/top/

\end{document}